%% file: main.tex
\documentclass[journal]{IEEEtran} 

\IEEEoverridecommandlockouts                              

\usepackage{cuted} 
\usepackage{etoolbox}

\input{preamble}

\usepackage{float}
\title{RoboVista: Evaluating Vision Language Models \\ for Diverse Robot Applications}

\author{

Shuangyu Xie$^{1,*}$, Kaiyuan Chen$^{1,*}$, Ziyang Chen$^{1}$,
Simeon Adebola$^{1}$, Yixuan Huang$^{2}$, Zehan Ma$^{1}$,\\

Tianshuang Qiu$^{1}$, Wentao Yuan$^{3}$, Dhruv Shah$^{2,3}$,
Pannag R. Sanketi$^{3}$, Ken Goldberg$^{1}$\\

$^{1}$University of California, Berkeley \quad
$^{2}$Princeton University \quad
$^{3}$Google DeepMind\\
$^{*}$Equal contribution

}

\begin{document}

\twocolumn[{%
\renewcommand\twocolumn[1][]{#1}%
\maketitle
\vspace{-20pt}
\begin{center}
    \centering
    \captionsetup{type=figure}
    \includegraphics[width=\linewidth]{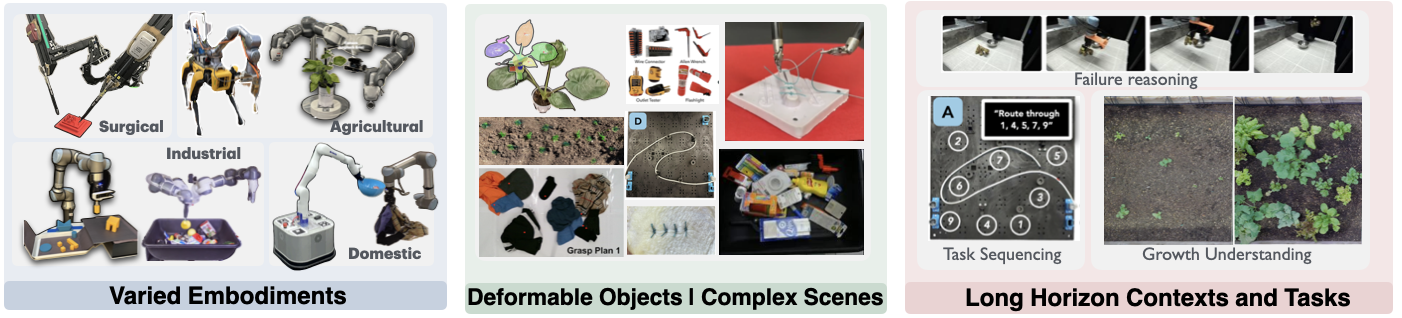}
    \captionof{figure}{\textbf{RoboVista Overview.} 
To support future robot applications, \benchname presents fine-grained spatial understanding and embodied decision-making challenges for Vision–Language Models (VLMs).
Grounded in 6 robot application domains and 39 diverse tasks, \benchname is an expert-annotated Visual Question Answering (VQA) dataset emphasizing variable robot embodiments (left), interactions with deformable objects and complex and cluttered scenes (middle), and long-horizon contextual understanding (right). }
    \label{fig:scene-title}
\end{center}%
}]

\begin{abstract}
\input{chapters/abstract}

\end{abstract}

\input{chapters/introduction}

\input{chapters/related_work}
\input{chapters/method}

\input{chapters/robovista}

\input{chapters/evaluation}

\input{chapters/physical}

\input{chapters/conclusion}

\renewcommand*{\bibfont}{\small}
\printbibliography
\newpage
\appendix
\input{chapters/appendix}

\end{document}

%% file: preamble.tex
\usepackage{microtype}
\microtypesetup{tracking=false}
\usepackage{amsmath} 
\usepackage{amssymb}  
\usepackage{amsfonts}
\usepackage{graphicx}
\usepackage[linesnumbered,ruled,vlined]{algorithm2e} 
\usepackage{algorithmic}
\usepackage{epsfig} 
\usepackage{xspace}
\usepackage{url}
\usepackage{comment}
\usepackage[dvipsnames]{xcolor}
\usepackage{colortbl}          
\usepackage[font=footnotesize]{caption}
\usepackage{booktabs} 
\usepackage{tikz}
\usetikzlibrary{positioning,shapes.geometric,arrows,fit,shapes.symbols,svg.path,tikzmark,calc}
\usepackage{textcomp}
\usepackage{listings}
\usepackage{subcaption}
\pgfdeclarelayer{background}
\pgfdeclarelayer{foreground}
\pgfsetlayers{background,main,foreground}
\usepackage{adjustbox}
\usepackage{array}
\usepackage{multirow}
\usepackage{dsfont}

\usepackage{siunitx}
\usepackage{mathtools}

\usepackage{physics}

\let\labelindent\relax 

\usepackage[inline]{enumitem} 

\usepackage{amsthm}

\usepackage{adjustbox} 
\usepackage{subcaption}
\usepackage{wrapfig}

\usepackage{multirow, makecell} 

\usepackage[dvipsnames,table]{xcolor}         
\usepackage{hyperref}       
\hypersetup{
    colorlinks=true,     
    urlcolor= WildStrawberry,
    citecolor=MidnightBlue,
    }

\definecolor{pregrasp}{RGB}{255,225,235}     
\definecolor{contact}{RGB}{255,245,220}      
\definecolor{grasp}{RGB}{230,250,230}        
\definecolor{release}{RGB}{220,235,255}      
\definecolor{postgrasp}{RGB}{235,220,245}     

\usepackage[backend=biber,
            hyperref=true,
            url=false,
            isbn=false,
            doi=false,
            backref=false,
            style=ieee,
            natbib=true,
            mincitenames=1,
            maxcitenames=10,   
            minbibnames=10,    
            maxbibnames=10,    
            citestyle=numeric-comp,
            sorting=none,
            block=none]{biblatex}
\renewcommand{\bibfont}{\small}
\newtheorem{Def}{Definition}

\usepackage{xspace}

\newcommand{\algname}{RQA\xspace}
\newcommand{\benchname}{RoboVista\xspace}

\definecolor{spatclr}{RGB}{215,226,244}   
\definecolor{goalclr}{RGB}{208,235,200}   
\definecolor{intclr} {RGB}{230,222,247}   

\usepackage{tabularx}
\usepackage{pifont} 
\usepackage[most]{tcolorbox}
\tcbset{colback=gray!10, colframe=gray!80!black, boxrule=0.5pt, arc=1mm, left=1mm, right=1mm, top=1mm, bottom=1mm, boxsep=1mm}

\definecolor{questionbg}{RGB}{240,248,255}
\definecolor{answergreen}{RGB}{34,139,34}
\definecolor{reasonblue}{RGB}{65,105,225}

\newtcolorbox{questionbox}{colback=questionbg,colframe=black!50,arc=4pt,auto outer arc,left=2mm,right=2mm,top=1mm,bottom=1mm}

\newcommand{\correctanswer}[1]{\textcolor{answergreen}{\textbf{Correct Answer:} #1}}

\newcommand{\rationale}[1]{\textcolor{reasonblue}{\textbf{Expert Rationale:} #1}}

\newcommand{\vqafigure}[5]{
\begin{figure}
    \centering
    \footnotesize
    \includegraphics[width=0.6\linewidth]{#1}
    \begin{questionbox}
        \textbf{Question:} #2
        \begin{itemize}[leftmargin=*]
            \item \correctanswer{#3}
            \item \rationale{#4}
        \end{itemize}
    \end{questionbox}
    \label{fig:#1}
\end{figure}
}

\newcommand{\vqafigurefixed}[5]{
\begin{figure}[H]
    \centering
    \footnotesize
    \includegraphics[width=0.6\linewidth]{#1}
    \begin{questionbox}
        \textbf{Question:} #2
        \begin{itemize}[leftmargin=*]
            \item \correctanswer{#3}
            \item \rationale{#4}
        \end{itemize}
    \end{questionbox}
    \label{fig:#1}
\end{figure}
}

\newcommand{\vqafigureduo}[5]{
\begin{figure}
    \centering
    \footnotesize
    \includegraphics[width=\linewidth]{#1}
    \begin{questionbox}
        \textbf{Question:} #2
        \begin{itemize}[leftmargin=*]
            \item \correctanswer{#3}
            \item \rationale{#4}
        \end{itemize}
    \end{questionbox}
    \label{fig:#1}
\end{figure}
}

%% file: chapters/abstract.tex
Diverse applications for robotics, such as industry and agriculture, require robots to operate across various embodiments, changing visual conditions, and complex planning. 
Vision–Language Models (VLMs) offer a promising foundation for general-purpose and interpretable robotic reasoning. 
Aligning VLMs with diverse robot applications requires a \textit{modular} understanding of the individual decision components that underlie robotic behavior. Capturing such structure is challenging for conventional robot benchmarks that are primarily based on teleoperated, end-to-end datasets. 
We propose Robot Question Answering (RQA), a modular evaluation framework and  \benchname, a benchmark curated from real robotic systems, research papers, and expert annotations. \benchname contains 474 Visual Question Answering (VQA) instances with human annotated reasoning and covers 39 unique task types in agricultural, industrial, domestic, surgical robotics, autonomous driving, and open robot datasets. 
Experiments on \benchname show that state-of-the-art VLMs exhibit substantial gaps.
Physical robot experiments suggest strong correlation between RoboVista performance and real-world task execution.~\url{https://berkeleyautomation.github.io/robovista}



%% file: chapters/introduction.tex


\section{Introduction}

Deploying robots in real-world domains such as industrial automation \cite{solowjow_indust_2020, adebola2024automating}, agriculture \cite{xie_energy_2025, adebolagarden2023}, and surgery \cite{chen2025surgical, haristitch22025} requires a general-purpose interface that can reason over diverse decisions, embodiments, and situations.
Vision-Language Models (VLMs)~\cite{radford2021learning, bordes2024introduction} are a potential backbone for these systems, as their language-driven reasoning has shown strong capabilities in complex domains such as software engineering \cite{si-etal-2025-design2code} and has shown early success in robot manipulation~\cite{geminiroboticsteam2025geminiroboticsbringingai, huang2023voxposer,ning2025prompting}. 
However, to support real robot tasks, VLMs must deliver consistent, spatially grounded decision-making: for example, on an industrial gasket assembly line, a VLM must accurately judge whether a gasket is properly seated and choose the correct next action (re-seat or fasten).
Toward this goal, we present \benchname, a high-quality benchmark that covers diverse robots application scenarios (as shown in Fig. ~\ref{fig:scene-title}). 


A primary way to evaluate this capability is Visual Question Answering (VQA)~\cite{antol2015vqa, kim_vqa2025, goyal2017making, lee2022what,yue_mmmu, yue-etal-2025-mmmu}, where a model is given images or videos and must answer task-relevant questions that probe scene understanding and action selection. 
Existing efforts, such as Robo2VLM~\cite{chenrobo2vlm} and RoboBrain~\cite{ji2025robobrain}, largely draw from imitation learning datasets such as Open X-Embodiment~\cite{open_x_embodiment_rt_x_2023}, and have shown success in improving spatial capabilities. 
However, many existing real-world robotic applications are fundamentally \emph{modular}: complex behaviors are decomposed into task-level decisions (e.g., task sequencing, action planning, and recovery) that are hard to capture by end-to-end robot trajectories. 
This motivates extending evaluation to these modular decisions with broader application domains, especially where public data is scarce, and many critical decisions are handled by analytical pipelines.


\begin{figure*}
    \centering
    \includegraphics[width=0.9\linewidth]{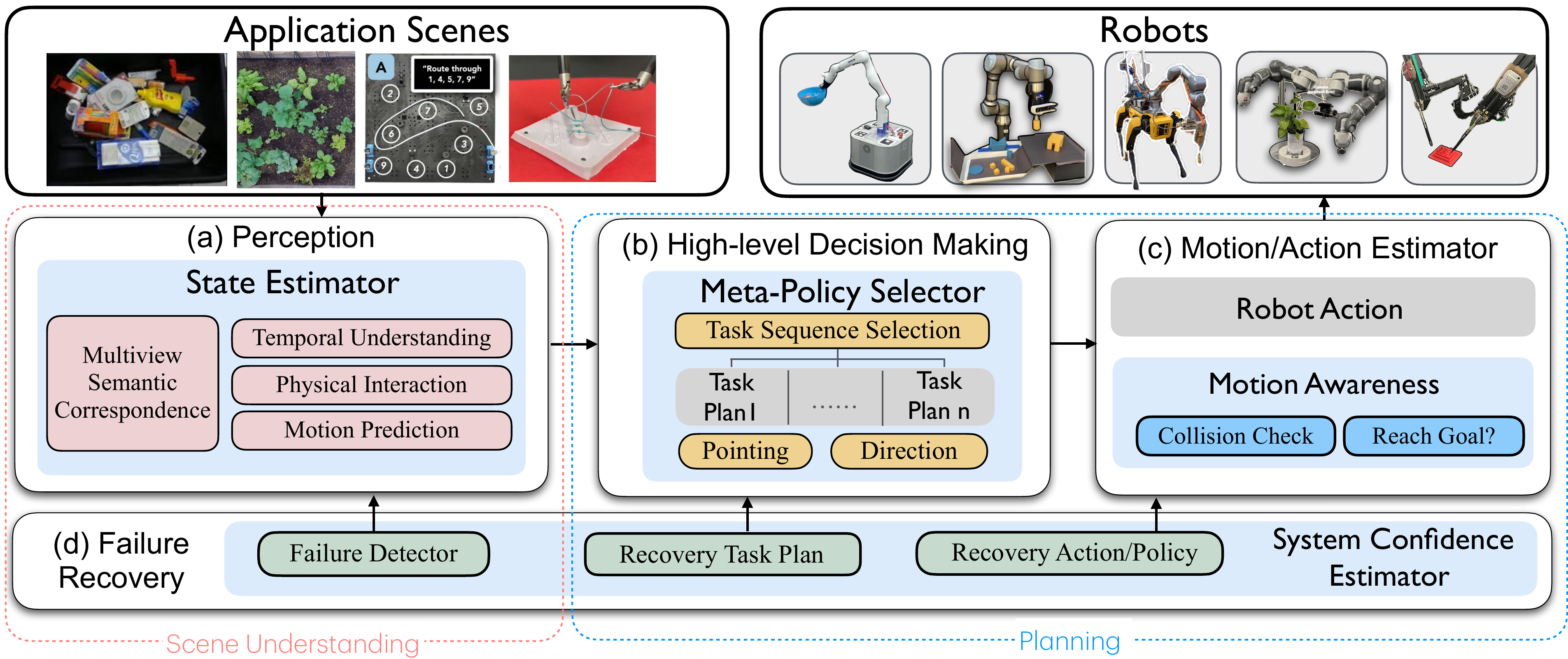}
    \caption{\textbf{Module abstraction of diverse robot applications pipelines}.
The figure illustrates a modular view of real-world robot systems, spanning diverse application scenes and robot embodiments. Robot operation can be decomposed into four functional layers in perception, high-level decision making, motion / action estimation and failure recovery. }
    \label{fig:task}
\end{figure*}



We propose \algname, a modular evaluation framework for Vision–Language Models (VLMs) that systematically constructs high-quality, robot-centric VQAs.
\algname decomposes diverse robot applications using standard robotic abstractions, and unifies human expert annotation, algorithmic task execution, and automated question construction within a shared Robot-VQA interface~\cite{yu2024manip}.
\algname provides a structured interface for evaluating VLM capabilities by formalizing both a module-level problem abstraction and a principled VQA construction process.

We introduce \benchname, an expert-annotated robot-centric VQA benchmark.
It contains 474 multiple-choice questions covering 39 distinct robot task types spanning surgical, agricultural, industrial, domestic, autonomous driving, and open robot datasets. Each question is grounded in robot-visible or onboard visual observations and paired with detailed human reasoning explanations.
To ensure realism and consistency across domains, all questions are constructed and verified by human domain experts familiar with the underlying robotic algorithms and task constraints.

We perform comprehensive evaluation of state-of-the-art VLMs on \benchname with varying techniques, such as Chain-of-Thought (CoT)~\cite{wei2022chain} and In-Context Learning (ICL)~\cite{dong2024survey, brown2020language}. 
Evaluation data suggests there is a substantial gap for VLM in many robot application domains.
We evaluate using a physical bi-manual spatial estimation task that uses domain-specific object geometry as priors.  We further evaluate closed-loop VLM-guided surgical knot tying. 
Evaluation data suggests that \benchname has strong correlation to task performance and the degradations in \benchname performance  significantly reduces long-horizon task success.

In this paper, we make the following contributions: 
(1) \algname: a modularized framework for constructing Visual Question Answering for VLMs from expert annotation, robot demonstration and execution;
(2) \benchname, a benchmark for VLMs spanning diverse robot applications, scenes, and embodiments; 
(3) a comprehensive evaluation of state-of-the-art VLMs on \benchname, including cross-domain analysis, training and complementary physical-robot validation.

%% file: chapters/related_work.tex
\section{Related Work}

\noindent\textbf{Modular Robot System Design.}
Modular robot system design \cite{yu2024manip} has been the dominant paradigm for field and service robotics. In this paradigm, complex robot behaviors are decomposed into well-defined functional tasks, such as perception, state estimation, motion reasoning, decision making, and control, that can be implemented and analyzed independently . This paradigm has enabled successful systems across a wide range of applications, including surgical robotics~\cite{chen2025surgical,surgical_debridement_dataset}, agricultural systems~\cite{xie2021toward,xie2023coupled,adebola2025botany}, industrial automation \cite{qiu2025omni,adebola2024automating}, and home robotics~\cite{adler2023teenager}. In many real-world domains, fully end-to-end data-driven approaches remain difficult to deploy due to the high cost and limited availability of training data. For example, surgical data must be obtained from real clinical procedures under strict safety, privacy, and regulatory requirements~\cite{chen2025surgical}. As a result, modularized pipelines continue to provide a practical and robust solution when data scaling itself is a hard problem, leveraging structured task decomposition and domain expertise.

\noindent\textbf{VLMs in Robotic Systems.}
Recent advances in Large Language Models (LLMs) and VLMs have demonstrated strong capabilities in task reasoning, spatial understanding, and grounded decision-making for robotic systems~\cite{Chen_2024_CVPR, karamcheti2024prismatic}. Increasingly, VLMs serve as the representational backbone of Vision–Language–Action (VLA) policies. Systems such as OpenVLA~\cite{kim2024openvla} and $\pi_0$\cite{pi0_2024} leverage pre-trained VLM representations to directly map visual observations and language goals to continuous robot actions. Beyond end-to-end control, VLMs are widely used for high-level task planning. Works such as Text2Motion\cite{lin2023text2motion} and PaLM-E~\cite{driess2023palme} integrate language or multimodal models into task-and-motion planning frameworks, using them as semantic planners that translate language instructions and multimodal observations into executable, long-horizon plans. Related approaches generate code or programmatic representations as intermediate planning interfaces~\cite{huang2023voxposer, huang2024rekep, fang2024moka, tang2025kalie}, enabling compositional reasoning over perception, control, and symbolic constraints. More recent systems adopt modularized architectures in which VLMs orchestrate or invoke specialized perception, planning, and control components~\cite{shi2025maestro}, combining the flexibility of foundation models with the structure of classical robotics pipelines. Despite this progress, it remains unclear whether the capabilities demonstrated by current VLMs are sufficient to support the full spectrum of robotic decision-making in real-world settings. Systematic evaluation is therefore required to reveal these gaps.

\noindent\textbf{VQA as an Evaluation Paradigm.}
VQA has emerged as a primary interface for evaluating and benchmarking the capabilities of VLMs~\cite{antol2015vqa, goyal2017making, lee2022what}. Early VQA benchmarks primarily assess general visual and linguistic abilities, such as commonsense knowledge (MMMU, MMMU-Pro~\cite{yue_mmmu, yue-etal-2025-mmmu}), chart and diagram understanding (ChartQA~\cite{masry-etal-2022-chartqa}), and mathematical reasoning (MathVista~\cite{lu2024mathvistaevaluatingmathematicalreasoning}). 

Recent embodied VQA benchmarks evaluate VLMs on tasks like visual navigation and long-horizon planning~\cite{das2018embodiedqa, anderson2018vision, zhao2025manipbench}, with newer works like ERQA~\cite{geminiroboticsteam2025geminiroboticsbringingai}, Robo2VLM~\cite{chenrobo2vlm}, and RoboMind~\cite{wu2025robomind} incorporating richer embodiment and stronger task grounding. While frameworks such as Robo2VLM~\cite{chenrobo2vlm} and RoboBrain~\cite{ji2025robobrain} leverage large-scale robot demonstrations to generate massive question-answer datasets for model post-training , RoboVista serves a complementary purpose. By designing VQA to capture perception-planning-control procedural methods across diverse applications , we enable evaluation in underrepresented domains—such as agriculture, surgery, and industrial robotics—that natively lack the massive trajectory corpora required by prior efforts. 

Meanwhile, the focus of RoboVista on fine-grained diagnostic quality aligns with a broader shift in the language modeling community toward small but challenging benchmarks, such as TurnaboutLLM \cite{yuan-etal-2025-turnaboutllm}, MuSR \cite{sprague2024musr}, and FrontierMath \cite{glazer2025frontiermathbenchmarkevaluatingadvanced}, that prioritize reasoning depth over raw scale. We consider our scale a deliberate design choice rather than a limitation by ensuring equal weight during evaluation and avoid over-representing data-rich tabletop scenarios. Rather than removing domain expertise, RoboVista provides a structured framework to integrate critical, modular decision points into robotic VQA.

%% file: chapters/method.tex
\section{RQA Framework}
We propose Robot Question Answering (RQA), a modular and unified framework that converts decision points from robot systems into a robot-centric VQA formulation. We first introduce a module-level abstraction of robot systems, then define the Robot-VQA data structure, describe the RQA construction process, and finally illustrate the framework with concrete case studies.

\subsection{Module Abstraction}

As illustrated in Fig.~\ref{fig:task}, RQA mirrors modular robot system pipelines~\cite{yu2024manip}, decomposing robot behavior into perception, decision making, motion execution, and recovery. 
Specifically, perception modules estimate task-relevant state from robot-centric visual observations, including object geometry, spatial relationships, and physical interactions. High-level decision making focuses on grounded task reasoning, requiring models to select goals, sequence actions, and determine appropriate next steps under task constraints. Motion and action awareness evaluates whether candidate actions are physically feasible, accounting for reachability, collision avoidance, and closed-loop visual feedback during execution. To handle failures that arise across different modules, recovery and robustness reasoning assess the ability to detect execution failures and infer corrective actions based on observed outcomes. Together, these stages provide a structured yet flexible abstraction for expressing diverse robot reasoning problems within a unified Robot-VQA formulation.


To facilitate the construction of Robot-VQA, we build upon established formalisms in perception~\cite{siciliano2016robotics}, planning \cite{lavalle2006planning, 10.5555/3165183}, and operational control \cite{khatib1987unified}. While individual abstractions exist for specific sub-problems, the general nature of Robot-VQA requires a unified framework. Inspired by the compositional logic of Neural Module Networks \cite{andreas2016neural}, we design a module-level problem abstraction where any functional block in the pipeline (Fig.~\ref{fig:task}) can be described by a tuple $\mathcal{M}$ with detailed explanations
\begin{equation}
    \mathcal{M} = (\mathcal{E}, \mathcal{X}, \mathcal{U}, \mathcal{C}),
\end{equation}
\begin{itemize}[leftmargin=1em]
    \item[$\mathcal{E}$] \textbf{Embodiment Setup} defines the robot's physical configuration and sensing capabilities. Following the configuration space approach \cite{lozano1983spatial}, this includes kinematic structures, actuation limits, and sensor modalities that dictate the robot's interaction with the environment.

    \item[$\mathcal{X}$] \textbf{State Space} represents the latent and observed variables of the robot and environment. Based on probabilistic robotics frameworks \cite{thrun2005probabilistic}, this includes joint configurations, object poses, and task-relevant semantic attributes necessary for state estimation.

\item[$\mathcal{U}$] \textbf{Task Output Space} defines the decision or control manifold, such as discrete actions or motion parameters. 
The module implements produce high-level task decisions or operational space commands \cite{khatib1987unified}.

\item[$\mathcal{C}$] \textbf{Constraints} encode feasibility conditions (e.g., collision avoidance, joint limits) that restrict the valid output space and enforce geometric and kinematic safety requirements~\cite{ratliff2009chomp}.
\end{itemize}




\subsection{Robot-VQA Data Structure}
Robot-VQA inherits from standard VQA while imposing robot-centric constraints to ensure embodied grounding.

\begin{Def}{(VQA Instance with Reasoning)}
\label{def:vqa}
Let $\mathbb{V}$, $\mathbb{Q}$, $\mathbb{A}$, and $\mathbb{R}$ denote the spaces of visual inputs, questions, answers, and textual rationales, respectively.  
A VQA instance with reasoning is a 5-tuple
$\mathcal{Q} = (\mathcal{V}, q, a^\star, \mathcal{A}, r)
\in \mathbb{V} \times \mathbb{Q} \times \mathbb{A} \times 2^{\mathbb{A}} \times \mathbb{R},$
where $\mathcal{V}$ is the visual input, $q$ the question, $a^\star$ the ground-truth answer, $\mathcal{A}$ the candidate answer set containing $a^\star$, and $r$ a textual rationale justifying $a^\star$.  
We denote the set of all such instances by $\mathbb{S}_{\text{VQA}}$.
\end{Def}

\begin{Def}{(Robot-Centric Visual Input)}
\label{def:rvqa-visual}
We define two robot-centric subsets of $\mathbb{V}$ such that robot is visible or image from onboard camera. 
\begin{align*}
    \mathbb{V}_{\text{rv}} := \{\mathcal{V} \in \mathbb{V} : \mathrm{RobVis}(\mathcal{V})\},
    \mathbb{V}_{\text{ob}} := \{\mathcal{V} \in \mathbb{V} : \mathrm{Onboard}(\mathcal{V})\}.
\end{align*}
\end{Def}

\begin{Def}{(Robot-VQA Instance)}
\label{def:robot-vqa}
A {Robot-VQA instance} is a subset of $\mathbb{S}_{\text{VQA}}$ whose visual input is grounded in the robot's embodied experience:
\[
\mathbb{S}_{\text{Robot-VQA}} := \left\{\mathcal{Q} = (\mathcal{V}, q, a^\star, \mathcal{A}, r) \in \mathbb{S}_{\text{VQA}} : \mathcal{V} \in \mathbb{V}_{\text{rv}} \cup \mathbb{V}_{\text{ob}}\right\}.
\]
\end{Def}

\subsection{\algname Design}

With the formal definition of module abstraction $\mathcal{M}$ and Robot-VQA instance
 $\mathbb{S}_{\text{Robot-VQA}}$, the key question of RQA framework is how to achieve the conversion from $\mathcal{M}$ to
 $\mathbb{S}_{\text{Robot-VQA}}$.
Concretely, RQA defines a structured mapping from a module specification to a Robot-VQA representation:
\begin{equation}
    (\mathcal{E}, \mathcal{X}, \mathcal{U}, \mathcal{C})
\;\longrightarrow\;
(\mathcal{V}, q, a^\star, \mathcal{A}, r),
\end{equation}
where the module’s latent state and embodiment are reflected through embodied visual observations, and its decision semantics are expressed through structured question–answering.

However, achieving robot-centric, modular VQA evaluation presents the following challenges for design considerations: 
(1) diverse robot applications rely on domain-specific modular pipelines, making it infeasible to provide a generic or fully automated construction procedure;
(2) many decisions further depend on embodiment-specific or non-visual state (e.g., kinematics, contact, reachability) and sometimes visual data is subject to occlusions, limited viewpoints, and clutter;
(3) designing plausible but incorrect answer choices is challenging, as each option must be algorithmically grounded and reflect a feasible outcome of the underlying module; 
(4) visual data and questions in certain domains and tasks are substantially harder to curate than in others.

 \begin{figure*}
    \centering
    \includegraphics[width=\linewidth]{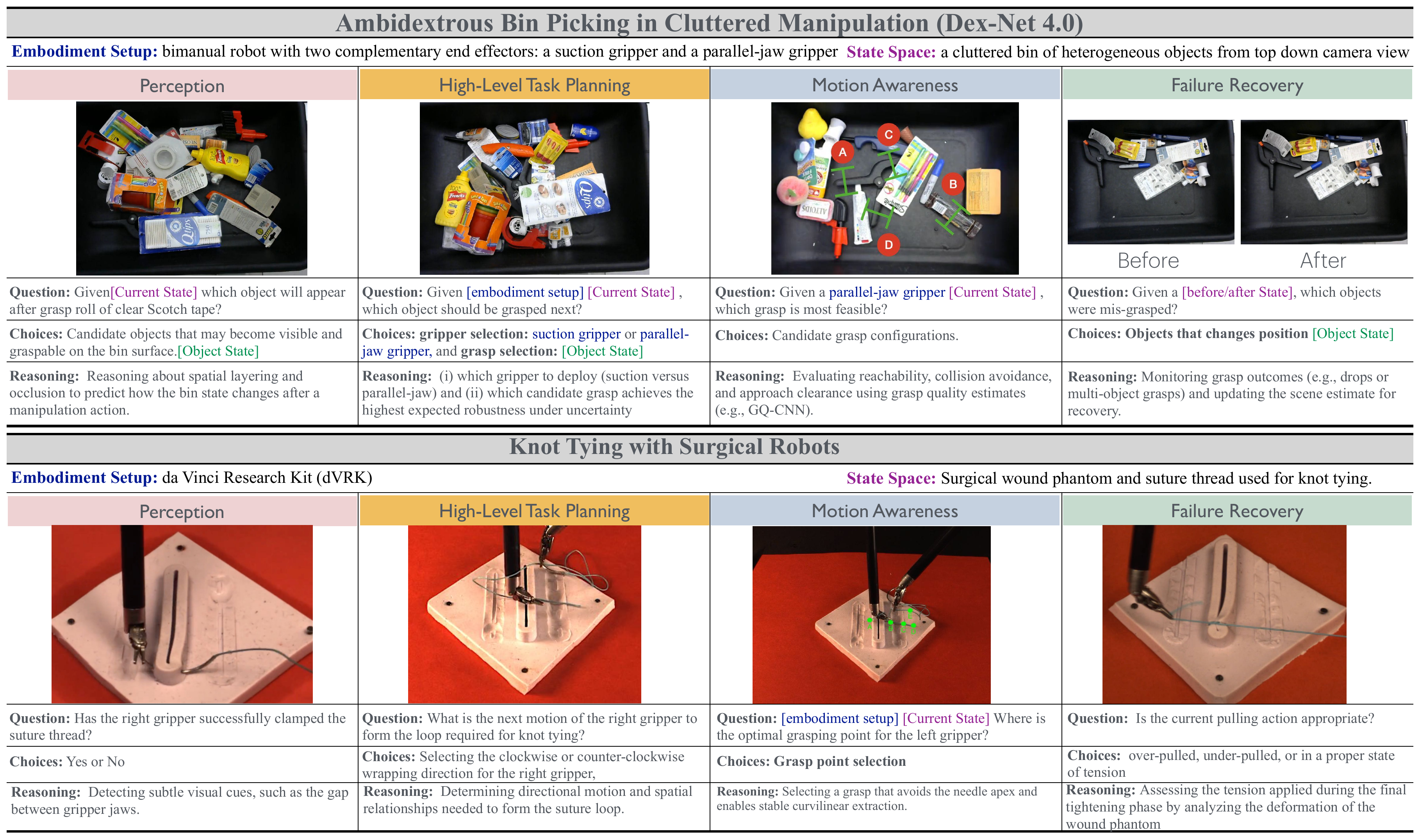}
    \caption{ \textbf{RQA case studies across domains.} This figure provide two example on how the RQA framework maps continuous physical robot states into modular, discrete VQA pairs that correspond to critical real-world decision points, such as tool alignment and tension management.
Top: Ambidextrous bin picking with Dex-Net~\cite{mahler2019learning} Bottom: Surgical knot tying~\cite{chen2025surgical}. In both cases, robotic decision points are decomposed into perception, high-level planning, motion awareness, and failure recovery with RQA.}
    \label{fig:RQA-exps}
    \vspace{-10pt}
\end{figure*}

To address these challenges, \algname adheres to the following design principles, each directly aligned with the challenges above:
(1) robot VQAs are decomposed and structured by domain experts familiar with the robot application system;
(2) each question is designed to be answerable with provided robot-centric visual and language context alone by the domain expert;
(3) all questions and answer choices admit a single, well-defined correct answer that can be verified against the original module outcome; 
(4) the question curation procedure intentionally limits redundancy within each question and task type and prioritizes coverage across diverse robot tasks and questions.

Under this formulation, RQA convert modules to Robot-VQA as following.
The visual input $\mathcal{V}$ consists of robot-centric images or image sequences that capture the task-relevant scene and robot configuration. These observations implicitly encode the underlying system state $\mathcal{X}$ used by the module, without explicitly exposing structured state variables in the VQA formulation.
The question $q$ describes the task-level decision associated with the module and specifies the objective to be reasoned about. When necessary, it also provides additional context about the embodiment setup $\mathcal{E}$ or task conditions that may not be visually explicit.
The ground-truth answer $a^\star$ corresponds to a valid module output, i.e., an element of the task output space $\mathcal{U}$. For evaluation, the output space is discretized into a finite candidate set $\mathcal{A} \subseteq \mathbb{A}$ that includes $a^\star$ as well as plausible alternative outputs.
Finally, embodiment, geometric, and kinematic constraints $\mathcal{C}$ governing the feasibility of module outputs are either observable from the visual input $\mathcal{V}$ or explicitly stated in the question $q$. The reasoning $r$ provides a textual justification explaining why $a^\star$ satisfies these constraints while alternative candidates do not.



\subsection{Case Studies: Instantiating RQA Across Domains}
We present two representative case studies for applying RQA designs: (1) ambidextrous bin picking and (2) surgical knot tying, to illustrate how Robot Question Answering (RQA) systematically exposes decision points from diverse robotic pipelines as Robot-VQA instances. Despite differences in embodiment, contact dynamics, and task horizons, both systems can be decomposed into a shared set of functional modules aligned with Fig.~\ref{fig:task}.

\noindent\textbf{Robot Grasping with Ambidextrous Robot}
Dex-Net 4.0~\cite{mahler2019learning} is a representative work for addressing the ambidextrous bin-picking problem, where a bimanual robot clears a cluttered bin of heterogeneous objects using two complementary end-effectors: a suction gripper and a parallel-jaw gripper. The system samples a large set of candidate grasps over the visible object surfaces and evaluates each candidate using a Grasp Quality Convolutional Neural Network (GQ-CNN), which predicts a grasp robustness score under uncertainty. The robot then selects both the gripper type and grasp configuration that jointly maximize expected grasp success. A set of example Robot-VQA questions are shown in Fig.~\ref{fig:RQA-exps} (top).
\input{tables/difficulty_table}
\noindent\textbf{Knot Tying with Surgery Robots}
Surgical knot tying is a long-horizon, contact-rich manipulation task that requires precise spatial reasoning, sequential decision making, fine-grained motion awareness, and continuous monitoring of execution outcomes. We frame knot tying on the da Vinci Research Kit (dVRK) \cite{kazanzides_dvrk} as a sequence of perception-grounded decision points. A set of example Robot-VQA questions are shown in Fig.~\ref{fig:RQA-exps} (bottom).\label{sec:rqa:knot}

%% file: tables/difficulty_table.tex
\begin{figure*}[t]
    \centering
    \begin{minipage}[t]{0.62\textwidth}
        \centering
        \vspace{2pt} 
        \includegraphics[width=\linewidth]{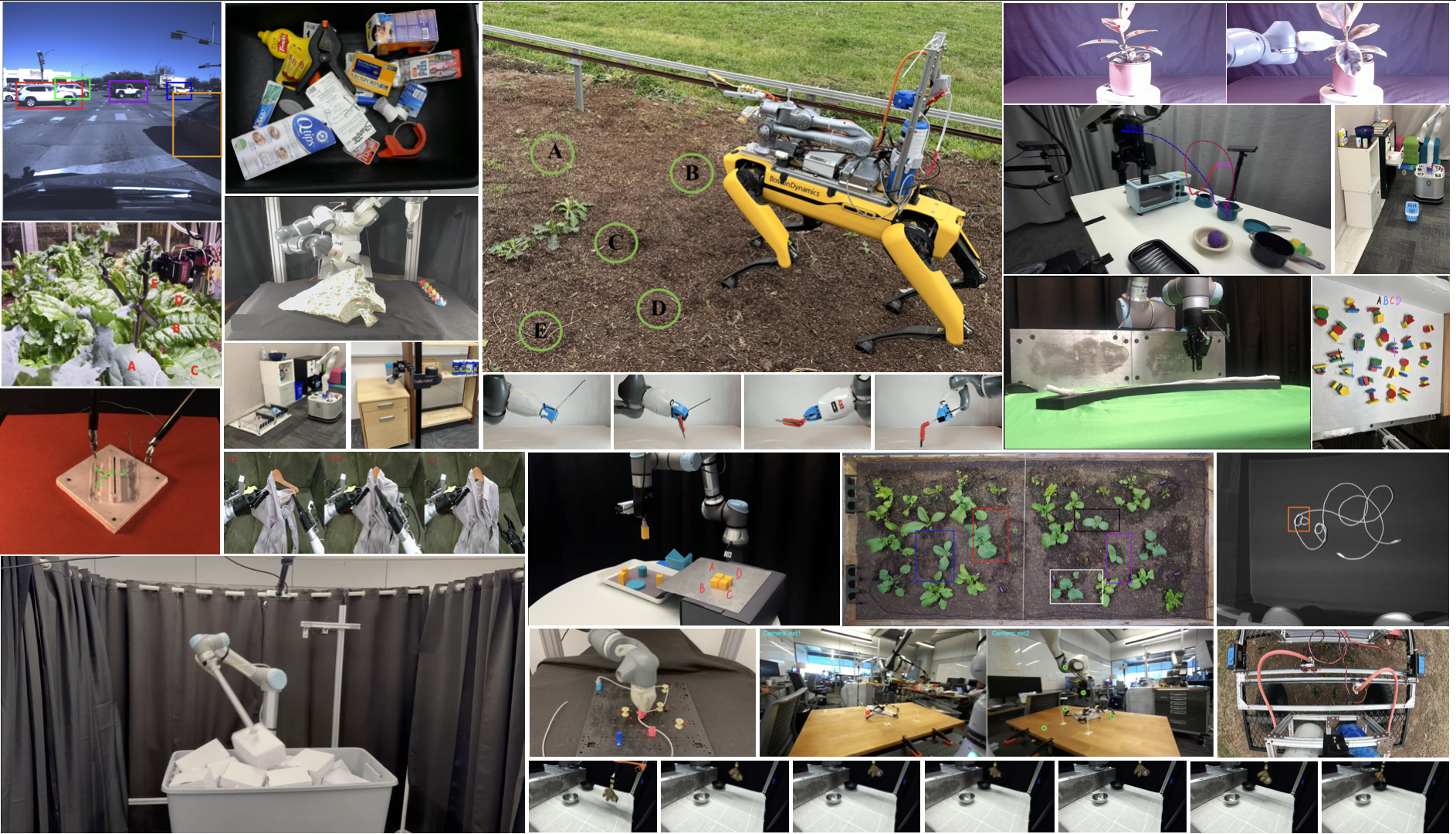}
        \label{fig:your_image}
    \end{minipage}
    \hfill
    \begin{minipage}[t]{0.36\textwidth}
        \captionsetup{type=table}
        \vspace{3pt}
        \setlength{\tabcolsep}{3pt}
        \footnotesize
        \resizebox{\linewidth}{!}{%
        \begin{tabular}{p{1.8cm} l | cc | c}
        \toprule
        \textbf{Domain} & \textbf{Task Description} & \textbf{Perception} & \textbf{Planning} & \textbf{Total} \\
        \midrule
        \cellcolor{green!15}Agriculture & Robot Gardening \cite{presten2022automated}& 18 & 0 & 18 \\
        \cellcolor{green!15} & Plant Inspection~\cite{adebola2025botany} & 12 & 4 & 16 \\
        \cellcolor{green!15} & Weed removal  \cite{weed_sparying, xie2023coupled, wang2024toward}& 19 & 9 & 28 \\
        \midrule
        \cellcolor{blue!15}Driving & Self-Driving in Mcity~\cite{gm_sae_autodrive_challenge_2021} & 9 & 11 & 20 \\
        \midrule
        \cellcolor{yellow!15}Domestic & Garment manipulation \cite{adler2023teenager} & 1 & 2 & 3 \\
        \cellcolor{yellow!15} &  Domestic Tidying \cite{huang2025points2plans, Huang-2025-fail2progress}& 30 & 19 & 49 \\
        \midrule
        \cellcolor{gray!15}Industrial & 1D deformable ~\cite{viswanath2023handloom, adebola2024automating} & 39 & 18 & 57 \\
        \cellcolor{gray!15} & 3D deformable~\cite{chen2023bagging, chen2022autobag}& 7 & 8 & 15 \\
        \cellcolor{gray!15} & Assembly \cite{goldberg2025blox} & 10 & 9 & 19 \\
        \cellcolor{gray!15} & Bin picking~\cite{tam2024bomp, mahler2019learning, agboh2023learning}& 26 & 17 & 43 \\
        \cellcolor{gray!15} & Defect scanning~\cite{qiu2025omni}& 10 & 0 & 10 \\
        \midrule
        \cellcolor{red!15}Surgical &  Knot Tying ~\cite{chen2025surgical}& 19 & 11 & 30 \\
        \cellcolor{red!15} & Debridement~\cite{surgical_debridement_dataset} & 11 & 5 & 16 \\
        \midrule
        \cellcolor{orange!15}Open Datasets & Agibot~\cite{bu2025agibot_iros} & 10 & 4 & 14 \\
        \cellcolor{orange!15} & DROID~\cite{droid} & 52 & 17 & 69 \\
        \cellcolor{orange!15} & Open X-Embodiment~\cite{open_x_embodiment_rt_x_2023} & 45 & 22 & 67 \\
        \midrule
        \textbf{Total} & & \textbf{318} & \textbf{156} & \textbf{474} \\
        \bottomrule
        \end{tabular}}
        
        \label{tab:difficulty}
    \end{minipage}
    \captionof{figure}{\textbf{\benchname Gallery}. Representative visual observations and tasks across the six domains, including agriculture, autonomous driving, domestic, industrial, surgical, and open robot datasets. The table summarizes task types and the distribution of perception- and planning-centric questions.}
    \vspace{-10pt}
\end{figure*}

%% file: chapters/robovista.tex
\section{The \benchname Benchmark}

We introduce \emph{\benchname}, a high quality robot-centric benchmark for VLMs constructed by RQA framework. \benchname consists of 474 expert-annotated, multiple-choice Robot-VQA questions, each paired with 5 candidate answers and detailed reasoning explanations annotated by human expert. All questions are grounded in robot-centric visual observations and are curated to reflect decision points from real robotic applications.
The visual data of \benchname is curated from six application domains, industrial manufacturing, agriculture, domestic robotics, surgical robotics, autonomous driving and open robot datasets.
It covers 39 distinct robotics task types, including bin picking, surgical knot tying, weed removal, cable routing, assembly, and navigation from 18 peer-reviewed conference and journal publications. 
This also includes questions from open robot datasets, DROID~\cite{droid}, Open X-Embodiment~\cite{open_x_embodiment_rt_x_2023} and Agibot~\cite{bu2025agibot_iros} with questions based on Robo2VLM framework~\cite{chenrobo2vlm}. 
To ensure annotation quality and task realism, all questions are annotated and verified by domain experts who are familiar with the underlying robotic algorithms and executions. All annotators are graduate-level and above, with more than half of the annotators holding Ph.D. degrees in Robotics.

\noindent The construction of \benchname follows a three-stage data curation pipeline designed to ensure task realism, visual grounding, and cross-domain consistency. 

\noindent\textbf{Data Collection.}
First, we survey a broad range of robotics datasets and peer-reviewed publications across industrial, agricultural, domestic, surgical, and driving domains. From these sources, we identify representative robot application scenarios that expose meaningful perception, decision-making, motion, and recovery challenges. We then extract robot-centric visual data, including single images and short image sequences captured from onboard or robot-visible viewpoints, and group them into candidate task episodes.

\noindent\textbf{Question Construction.}
In the second stage, we invite human domain experts (four postdoctoral researchers and three students with master degree) to manually construct VQA questions from the selected episodes. 
Each question is designed to correspond to a concrete decision point within a modular robotics pipeline, such as grasp selection, next-action prediction, motion feasibility checking, or failure diagnosis. All questions are formulated as multiple-choice with five options, ensuring that each correct answer is visually grounded and operationally meaningful for robot execution. Annotators also provide a natural language explanation to justify the correct choice and clarify task intent.

\noindent\textbf{Quality Control.}
To ensure high annotation quality and consistency, all questions undergo a multi-pass review process. Each VQA instance is independently verified by at least one additional annotator with robotics expertise, who checks visual grounding, answer correctness, and linguistic ambiguity. Questions that admit multiple plausible answers, rely on non-visual cues, or lack clear task relevance are revised or discarded. We further enforce consistency across domains by aligning question types and difficulties. This process results in a curated benchmark that balances coverage, difficulty, and realism while minimizing annotation noise.


%% file: chapters/evaluation.tex
\section{Experiments}
In this section, we benchmark state-of-the-art open-source models in different configurations, including different sizes of Qwen2.5-VL~\cite{bai2025qwen2} and Qwen3-VL~\cite{yang2025qwen3}. 
For VLMs that are trained for robotics data, we benchmark Robo2VLM-ER~\cite{chenrobo2vlm}, RoboBrain 2.5~\cite{tan2026robobrain}. For RoboBrain 2.5, we include a generic model for VQA and a model for spatial understanding (NV). Robo2VLM-ER uses Qwen2.5-VL-7B as base model, and  RoboBrain 2.5 uses Qwen3-VL-8B as base model.
For closed-source VLMs, we benchmark GPT-4o, GPT-5 (gpt-5-2025-08-07) and Gemini 2.5 Pro~\cite{comanici2025gemini}. GPT-5 and  Gemini 2.5 Pro are released at similar time frame.
We also include results from Qwen-3-8B without any visual input to show how much linguistic hints provided in choice designs.
All models are evaluated with a temperature of 0.7, a maximum completion token length of 4096, and overall context length of 10240.

\subsection{Results on Zero-Shot Performance}

\noindent\textbf{Differences in VLM Models}
Table~\ref{tab:zero-shot-performance} summarizes zero-shot accuracy across all questions and across six domains.
\benchname overall poses non-trivial challenges. Even the best-performing models fall substantially short of perfect performance. Closed-source models achieve the strongest aggregate performance, with Gemini 2.5 Pro obtaining the highest overall accuracy (56.5\%).
Among open-source models, performance improves consistently with model scale, with Qwen 3-235B-A22B reaching 51.3\% overall accuracy. Notably, robotics-specialized models such as RoboBrain 2.5-8B outperform similarly sized general-purpose VLMs on several domains, such as Driving and Surgery. However, these gains are not uniform across domains: for example, RoboBrain 2.5-8B-NV improves performance in domestic and surgical robotics tasks but underperforms in agricultural scenes. 

\noindent\textbf{Differences in Application Domains}  Across domains, performance varies substantially depending on the underlying visual properties and question difficulties. 
Domestic robotics consistently yields higher accuracies for most models, as these scenes are typically well-lit, human-scale, and visually structured. 
In contrast, agricultural robotics remains one of the most challenging domains.
Agricultural scenes often involve fine-grained plant morphology, severe self-occlusion, and large intra-class appearance variation across growth stages, species, and environmental conditions.
Questions in this domain frequently require reasoning about deformable, partially observable structures (e.g., leaves, stems, weeds) and subtle geometric cues that are weakly represented in existing VLM pretraining corpora.
These factors lead to consistently lower performance, even for large-scale and robotics-specialized models.
\input{tables/model_performance}

\subsection{Results on Chain-of-Thoughts Prompting}
\input{tables/cot}

Chain-of-Thought (CoT) prompting~\cite{wei2022chain} is a widely used technique for eliciting explicit intermediate reasoning in large language models and has  shown performance increase in some robotics tasks~\cite{geminiroboticsteam2025geminiroboticsbringingai}.
We evaluate CoT by appending the following instruction to the end of each question:
\emph{``Think step by step about this question, then provide your final answer.''}
All other prompting and decoding settings are kept identical to the zero-shot evaluation.

\noindent\textbf{Effect on Scene Understanding.}
Table~\ref{tab:cot-integrated} shows the effect of CoT prompting on Scene Understanding questions, measured as the accuracy difference between CoT and zero-shot prompting.
Across most models and domains, CoT consistently reduces scene understanding accuracy, with overall drops of up to 12\%.
The degradation is most significant in domains such as Driving and Domestic robotics, where questions require precise and direct spatial perception.
This follows similar observation as an empirical study on VLM thinking~\cite{lithink} that enforcing explicit reasoning traces sometimes lead the models to overly attend to language semantics and can interfere with fine-grained perceptual processing.

\noindent\textbf{Effect on Planning.}
In contrast to its impact on perception, CoT generally improves planning performance for several models, particularly in multi-step domains such as Surgical and Agricultural robotics.
These tasks often require sequencing actions, reasoning over task constraints, or anticipating future outcomes, where explicit intermediate reasoning can be beneficial. 
However, completely isolating planning from perception is not possible, so balancing the right amount of thinking is important for the planning tasks with VLMs.

\subsection{Results on In-Context Learning}
\input{tables/icl_table}

In-context learning (ICL)~\cite{brown2020language} refers to conditioning a model on a small number of example question--answer pairs provided directly in the prompt, without any parameter updates.
ICL has been shown to improve performance in general VQA and reasoning benchmarks by implicitly adapting model behavior to the target task distribution.
We evaluate ICL by prepending a small set of representative Robot-VQA examples with explicit reasoning steps to the prompt. We provide a random example on the same domain and task, and average the accuracy across five queries. All other evaluation settings remain identical to the zero-shot baseline.

\noindent\textbf{In-Context Learning Effects}. 
Table~\ref{tab:icl-compact} summarizes the effect of ICL. 
Even providing in-context examples of the same domain and task, 
ICL consistently \emph{reduces} accuracy across all evaluated models, with drops ranging from 2.8\% to 6.5\%.
We observe the degradation for both general-purpose and robotics-specialized VLMs. 
These results suggest that, unlike abstract reasoning benchmarks, providing in-context examples about spatial relationship may distract VLMs in longer contexts and lead to hallucination. 
However, we also recognize that larger models with stronger reasoning capabilities can be less affected by the in-context examples.

\noindent\textbf{In-Context Learning can lead to hallucination}.
To further analyze the effect of ICL, we examine 
 hallucination with model calibration~\cite{wei2024measuring}.
To measure calibration, we prompt models to provide both an answer and their confidence
from 0\% to 100\% with the setup as Wei et al~\cite{wei2024measuring} and Humanity’s Last Exam~\cite{phan2025humanity}.
Intuitively, a well-calibrated model should express high confidence only when it is likely to be correct; systematic over-confidence indicates a tendency to hallucinate plausible but incorrect answers.
As shown in Table~\ref{tab:icl-compact}, ICL consistently \emph{increases} calibration error across all models, with absolute CE increases of up to 9.7\%.
This trend indicates that ICL encourages models to produce more confident but less reliable predictions, which could amplify hallucinated reasoning when visual evidence is ambiguous or weakly grounded.

\subsection{Results on Failure Analysis}

\begin{figure}
    \centering
    \includegraphics[width=\linewidth]{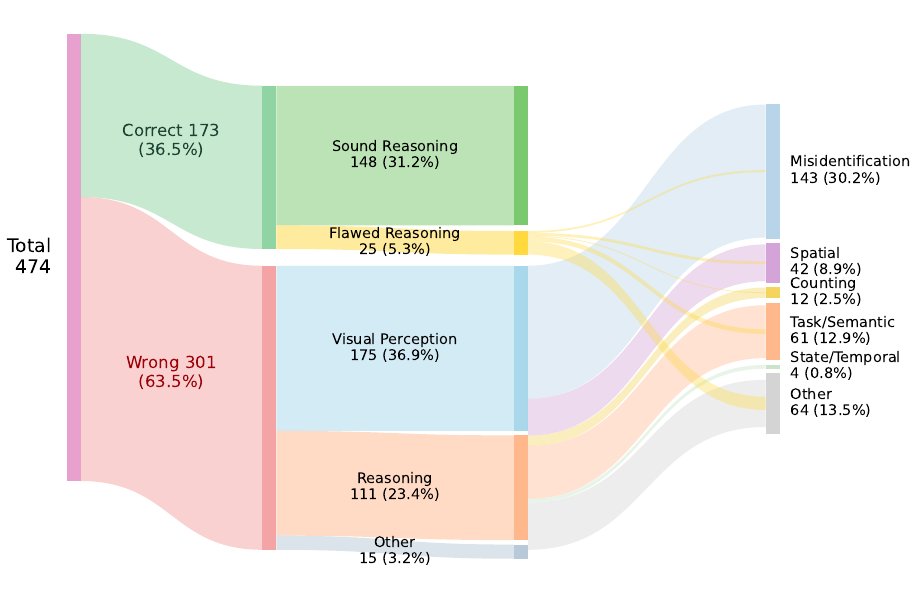}
    \caption{\textbf{Failure Analysis of Qwen2.5-VL 7B}. We analyze the reasoning chain of Qwen2.5-VL 7B and categorize failures in misidentification and reasoning.  }
    \label{fig:failure}
    \vspace{-10pt}
\end{figure}

\begin{figure}
    \centering
    \includegraphics[width=\linewidth]{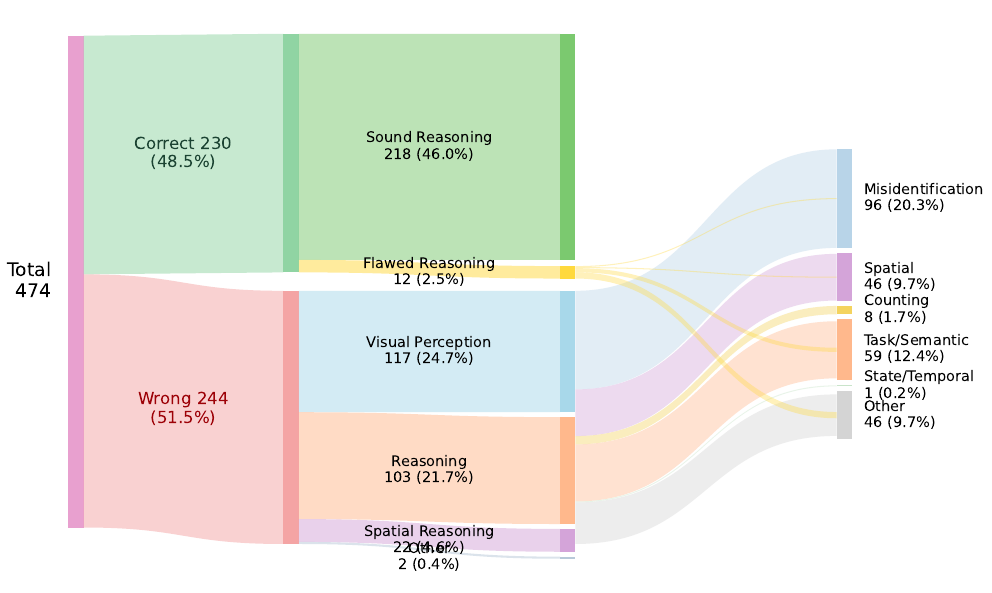}
    \caption{\textbf{Failure Analysis of Qwen3-VL 235B}. We analyze the reasoning chain of Qwen3-VL 235B and categorize failures in misidentification and reasoning.  }
    \label{fig:failure235b}
\end{figure}

We conduct failure analysis using model-generated reasoning chains that separates visual perception errors from higher-level reasoning failures.  For both correct and incorrect predictions, we use Qwen3-VL-32B-Thinking to propose an initial failure category given groundtruth human reasoning trace and answer. Then we manually verify and correct these assignments.

\noindent\textbf{Failure Analysis on Qwen2.5-VL-7B}. Figure~5 shows the failure breakdown. 
A key observation is that the majority of failures originate from visual perception rather than logical reasoning.
The most frequent failure mode is misidentification, accounting for 143 cases (30.2\%), primarily involving incorrect object identity, state, or spatial location.
Common examples include confusing visually similar objects, or incorrectly localizing target objects under occlusion or clutter.
Spatial reasoning errors form the second largest category, encompassing failures in relative positioning, depth estimation, motion inference, and reachability.
Task and semantic misinterpretation reflects failures in mapping visual evidence to task intent, such as misunderstanding which object is being manipulated or misinterpreting contact and interaction states.

\noindent\textbf{Failure Analysis on Qwen3-235B-A22B }.  We perform the same analysis for the much larger 235B model.
Scaling significantly reduces both the overall error rate and the proportion of flawed reasoning chains, with correct answers increasing from 36.5\% to 48.5\%.
Misidentification errors drop substantially (from 30.2\% to 20.3\%), indicating improved visual recognition and grounding at scale.
However, spatial and semantic errors persist, and remain a dominant source of failure even for the largest model.
This suggests that while increased model capacity improves visual robustness and reasoning consistency, fundamental challenges in spatial understanding and geometry-aware perception are not fully resolved by scale alone.

%% file: tables/model_performance.tex
\begin{table}
    \centering
    \caption{\textbf{Zero-Shot Performance of Multimodal Foundation Models on RQA (\%).}}
    \captionsetup{font=footnotesize}
    \setlength{\tabcolsep}{4pt}

    \sisetup{
        table-format=2.1,
        table-space-text-post={\%},
        output-decimal-marker={.},
        table-parse-only,
        input-symbols = {-- xx}
    }

    \resizebox{\linewidth}{!}{%
    \begin{tabular}{@{}l
        S[table-format=2.1]
        S[table-format=2.1]
        S[table-format=2.1]
        S[table-format=2.1]
        S[table-format=2.1]
        S[table-format=2.1]
        S[table-format=2.1]@{}}

    \toprule
    \textbf{Model} &
    {\textbf{All}} &
    {\cellcolor{green!15}\textbf{Agriculture}} &
    {\cellcolor{blue!15}\textbf{Driving}} &
    {\cellcolor{yellow!15}\textbf{Home}} &
    {\cellcolor{gray!15}\textbf{Industry}} &
    {\cellcolor{red!15}\textbf{Surgery}} &
    {\cellcolor{orange!15}\textbf{Open Datasets}} \\
    \midrule

    \multicolumn{8}{@{}l}{\textit{Baselines}} \\
    Random  & 20.0 & 20.0 & 20.0 & 20.0 & 20.0 & 20.0 & 20.0 \\
    Qwen 3-8B (Text-Only) & 25.1 & 27.4 & 30.0 & 30.8 & 22.2 & 26.1 & 24.0 \\  
    \midrule
    \multicolumn{8}{@{}l}{\textit{Closed-Source Models}} \\
    GPT-4o & 49.6 & \textbf{50.0} & 50.0 & 59.2 & 32.5 & 67.4 & 53.5 \\
    GPT-5 & 48.1 & 38.7 & \textbf{55.0} & 46.1 & 35.7 & 63.0 & \textbf{58.3} \\
    Gemini 2.5 Pro & \textbf{56.5} & 48.4 & 50.0 & \textbf{63.2} & \textbf{48.4} & \textbf{76.1} & \textbf{58.3} \\
    \midrule

    \multicolumn{8}{@{}l}{\textit{Open-Source Models}} \\
    Qwen 2.5 VL-3B & 39.9 & 33.9 & 40.0 & 43.4 & 27.8 & 50.0 & 47.9 \\
    Qwen 2.5 VL-7B & 43.7 & 37.1 & 45.0 & 47.4 & 32.5 & 52.2 & 51.4 \\
    Qwen 2.5 VL-32B & 43.9 & \textbf{48.4} & 50.0 & 46.1 & 31.0 & 58.7 & 46.5 \\
    Qwen 2.5 VL-72B & 44.3 & 43.5 & 35.0 & 40.8 & 31.7 & \textbf{69.6} & 50.7 \\
    Robo2VLM-ER & 42.6 & 35.5 & 40.0 & 43.4 & 32.5 & 54.3 & 50.7 \\
    RoboBrain 2.5-8B & 45.8 & 37.1 & 55.0 & 51.3 & 36.5 & 56.5 & 50.0 \\
    RoboBrain 2.5-8B-NV & 47.0 & 27.4 & 40.0 & \textbf{56.6} & 38.9 & 65.2 & 52.8 \\
    Qwen 3 VL-4B & 44.3 & 38.7 & 45.0 & 42.1 & 34.9 & 54.3 & 52.8 \\
    Qwen 3 VL-8B & 46.8 & 38.7 & 45.0 & \textbf{56.6} & 32.5 & 58.7 & 54.2 \\
    Qwen 3 VL-32B & 49.2 & \textbf{48.4} & 55.0 & 48.7 & 35.7 & 65.2 & 55.6 \\
    Qwen 3-235B-A22B & \textbf{51.3} & 46.8 & \textbf{60.0} & 53.9 & \textbf{37.3} & \textbf{69.6} & \textbf{56.9} \\

    \bottomrule
    \end{tabular}
    }
    \label{tab:zero-shot-performance}
\end{table}

%% file: tables/cot.tex
\begin{table}[t]
    \centering
    \caption{\textbf{Chain-of-Thought Effect on Perception and Planning.} Comparing zero-shot baseline vs CoT prompting. \textcolor{ForestGreen}{Green} = improvement, \textcolor{red}{Red} = degradation.  CoT consistently reduces scene understanding accuracy due to over-thinking but performs better with planning related tasks.}
    \setlength{\tabcolsep}{2pt}
    \scriptsize
    \begin{tabular}{@{}l ccc ccc ccc@{}}
    \toprule
    & \multicolumn{3}{c}{\textbf{Overall}} & \multicolumn{3}{c}{\textbf{Perception}} & \multicolumn{3}{c}{\textbf{Planning}} \\
    \cmidrule(lr){2-4} \cmidrule(lr){5-7} \cmidrule(lr){8-10}
    \textbf{Model} & Base & CoT & $\Delta$ & Base & CoT & $\Delta$ & Base & CoT & $\Delta$ \\
    \midrule
    Qwen 2.5 VL-3B & 39.9 & 29.5 & \textcolor{red}{-10.3} & 41.2 & 29.2 & \textcolor{red}{-11.9} & 37.2 & 30.1 & \textcolor{red}{-7.1} \\
    Qwen 2.5 VL-7B & 43.7 & 36.5 & \textcolor{red}{-7.2} & 45.9 & 34.6 & \textcolor{red}{-11.3} & 39.1 & 40.4 & \textcolor{ForestGreen}{+1.3} \\
    Qwen 2.5 VL-32B & 43.9 & 37.6 & \textcolor{red}{-6.3} & 44.7 & 38.4 & \textcolor{red}{-6.3} & 42.3 & 35.9 & \textcolor{red}{-6.4} \\
    Qwen 2.5 VL-72B & 44.3 & 43.0 & \textcolor{red}{-1.3} & 45.6 & 41.2 & \textcolor{red}{-4.4} & 41.7 & 46.8 & \textcolor{ForestGreen}{+5.1} \\
    Robo2VLM-ER & 42.6 & 37.6 & \textcolor{red}{-5.1} & 45.6 & 36.8 & \textcolor{red}{-8.8} & 36.5 & 39.1 & \textcolor{ForestGreen}{+2.6} \\
    RoboBrain 2.5-8B & 45.8 & 40.7 & \textcolor{red}{-5.1} & 47.2 & 40.3 & \textcolor{red}{-6.9} & 42.9 & 41.7 & \textcolor{red}{-1.3} \\
    RoboBrain 2.5-8B-NV & 47.0 & 41.6 & \textcolor{red}{-5.5} & 48.1 & 40.9 & \textcolor{red}{-7.2} & 44.9 & 42.9 & \textcolor{red}{-1.9} \\
    Qwen 3 VL-4B & 44.3 & 44.9 & \textcolor{ForestGreen}{+0.6} & 46.9 & 45.6 & \textcolor{red}{-1.3} & 39.1 & 43.6 & \textcolor{ForestGreen}{+4.5} \\
    Qwen 3 VL-8B & 46.8 & 44.9 & \textcolor{red}{-1.9} & 47.2 & 46.9 & \textcolor{red}{-0.3} & 46.2 & 41.0 & \textcolor{red}{-5.1} \\
    Qwen 3 VL-32B & 50.4 & 52.1 & \textcolor{ForestGreen}{+1.7} & 53.1 & 51.6 & \textcolor{red}{-1.6} & 44.9 & 53.2 & \textcolor{ForestGreen}{+8.3} \\
    GPT-5 & 55.5 & 55.7 & \textcolor{ForestGreen}{+0.2} & 56.6 & 56.3 & \textcolor{red}{-0.3} & 53.2 & 54.5 & \textcolor{ForestGreen}{+1.3} \\
    \bottomrule
    \end{tabular}
    \vspace{-10pt}
    \label{tab:cot-integrated}
\end{table}

%% file: tables/icl_table.tex

\begin{table}[t]
    \centering
    \caption{\textbf{In-Context Learning Effect.} Comparing zero-shot baseline vs ICL with reasoning examples.  Calibration Error (CE) measures overconfidence as a typical indicator for hallucination ~\cite{wei2024measuring} ($\downarrow$ = better). }
    \setlength{\tabcolsep}{3pt}
    \footnotesize
        \begin{tabular}{@{}l ccc ccc@{}}
        \toprule
        & \multicolumn{3}{c}{\textbf{Accuracy (\%)$\uparrow$}} & \multicolumn{3}{c}{\textbf{Calibration Error (\%)$\downarrow$}} \\
        \cmidrule(lr){2-4} \cmidrule(lr){5-7}
        \textbf{Model} & Base & ICL & $\Delta$ & Base & ICL & $\Delta$ \\
        \midrule
        Qwen 2.5 VL-3B & 39.7 & 35.8 & \textcolor{red}{-3.8} & 46.3 & 51.3 & \textcolor{red}{+5.0} \\
        Qwen 2.5 VL-7B & 43.0 & 39.1 & \textcolor{red}{-4.0} & 41.9 & 46.2 & \textcolor{red}{+4.4} \\
        Qwen 2.5 VL-32B & 42.0 & 39.2 & \textcolor{red}{-2.8} & 50.8 & 54.5 & \textcolor{red}{+3.7} \\
        RoboBrain-MT & 47.5 & 41.0 & \textcolor{red}{-6.5} & 38.7 & 48.4 & \textcolor{red}{+9.7} \\
        RoboBrain-NV & 47.3 & 42.7 & \textcolor{red}{-4.6} & 37.2 & 46.0 & \textcolor{red}{+8.8} \\
        Qwen 3 VL-4B & 43.0 & 39.3 & \textcolor{red}{-3.7} & 49.7 & 54.1 & \textcolor{red}{+4.4} \\
        Qwen 3 VL-8B & 47.3 & 42.3 & \textcolor{red}{-5.0} & 47.2 & 52.8 & \textcolor{red}{+5.6} \\
        Qwen 3 VL-32B & 48.3 & 46.6 & \textcolor{red}{-1.7} & 44.6 & 47.7 & \textcolor{red}{+3.1} \\
        \bottomrule
        \end{tabular}
    \label{tab:icl-compact}
    \vspace{-8pt}
\end{table}

%% file: chapters/physical.tex

\section{Physical Evaluation}
In this evaluation, we examine how VLMs perform on RoboVista benchmark correlates with two representative robotic tasks. Specifically, we study: (1) bimanual gripper position alignment, (2) surgical knot-tying.

\subsection{Bimanual Gripper Position Alignment}
In this experiment, we study how VLMs perform on 3D spatial understanding and symbolic planning tasks. 



\begin{figure}[t]
    \centering
    \begin{subfigure}[t]{0.9\linewidth}
        \centering
        \includegraphics[width=\linewidth]{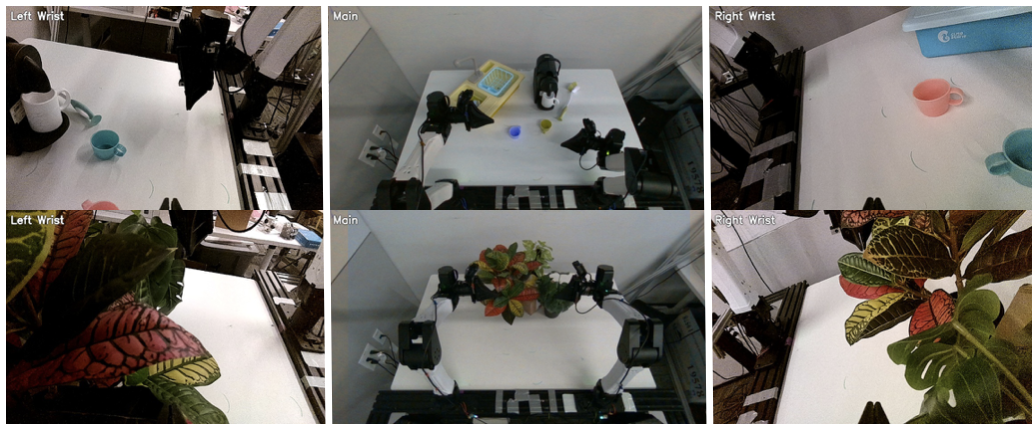}
        \caption{Two bi-manual gripper alignment setup examples.}
        \label{fig:bimanual-setup}
    \end{subfigure}

    \begin{subfigure}[t]{0.9\linewidth}
        \centering
        \includegraphics[width=\linewidth]{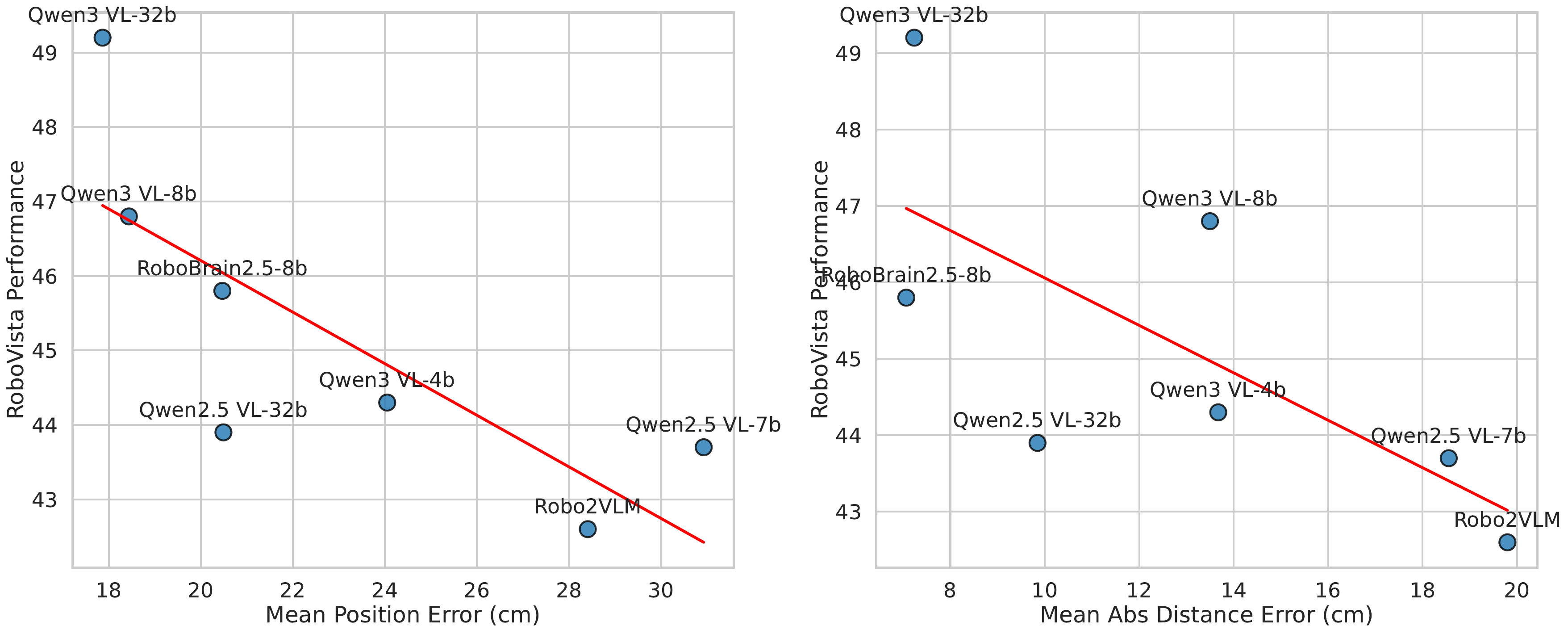}
        \caption{Correlation between RoboVista performance and (left) final gripper position error and (right) mean absolute distance estimation error.}
        \label{fig:bimanual-correlation}
    \end{subfigure}

    \caption{\textbf{Bi-manual gripper alignment task Setup and correlation with RoboVista performance.}
    The VLM estimates the distance between gripper tips and plans motions based on spatial priors. Higher RoboVista scores strongly correlates to lower estimation and execution errors.}
    \label{fig:bimanual}
    \vspace{-10pt}
\end{figure}

\noindent\textbf{Setup}.
As shown in Fig.~\ref{fig:bimanual-setup}, we use a bimanual manipulation task that requires VLMs to reason about spatial relationships and generate symbolic motion plans. Given visual observation from the main camera and two wrist cameras of the robot, the VLM is asked to (1) estimate the distance between the left and right gripper tips, and (2) generate a sequence of symbolic motion commands with associated numerical distances that move the right gripper to align with and touch the left gripper.
We integrate VLM as following: the visual input consists of images from the left wrist camera, right wrist camera, and a top-down main camera view, as illustrated in Fig.~\ref{fig:bimanual-setup}. The language query specifies the task objective and defines a robot-centric coordinate system, where forward/backward correspond to the $x$-axis, left/right to the $y$-axis, and up/down to the $z$-axis. The model is required to output a motion plan as a sequence of symbolic directional commands paired with metric distances, following a fixed format, e.g.,
$\{(\texttt{left},\,10\,\mathrm{cm}),\;(\texttt{up},\,5\,\mathrm{cm})\}$.
We evaluate the robot under five distinct geometrical priors: (1) kitchen objects, (2) small industrial parts, (3) no background object, (4) checkerboard (5) plants. For each prior, we set 5 configurations with the left and right arms randomly positioned to cover the workspace and to span inter-gripper distances ranging from 10 to 40,cm. Ground-truth distances are computed using the robot’s joint encoders and forward kinematics.


\noindent\textbf{Results}.
We evaluate multiple VLMs on the bi-manual alignment task and report results averaged across all robot configurations (Fig.~\ref{fig:bimanual-correlation}).  
Across both distance estimation error and the final position error after executing the predicted motion plan, we observe a strong negative correlation between benchmark performance and physical error: models with higher RoboVista scores consistently achieve lower distance estimation errors and smaller final alignment errors after execution. In particular, the position error exhibits a strong negative correlation with RoboVista performance (Pearson $r=-0.78$; Spearman $\rho=-0.93$), and distance error shows a moderate negative correlation (Pearson $r=-0.70$; Spearman $\rho=-0.75$). This indicates a strong monotonic relationship between benchmark accuracy and real-world task performance. 

\subsection{Surgical Knot-Tying using dVRK} 

To evaluate the relationship between benchmark performance and real-world task success, we perform integrated and closed-loop surgical knot tying using the dVRK as described in Section~\ref{sec:rqa:knot}.  This experiment evaluates whether performance measured by RoboVista in a surgical domain correlates with a model’s ability to support long-horizon, physically grounded task execution.

\noindent\textbf{Setup}.
We consider a shared-autonomy setting for surgical knot tying, motivated by scenarios in which a vision-language model assists junior operators during task execution. The physical setup is simulated to allow controlled evaluation while preserving the visual and procedural structure of real knot-tying workflows. Human experts with surgical robotics experience design a sequence of VQA questions with \algname framework that correspond to critical decision points in the knot-tying process, such as tool positioning, loop formation, tension management, and error detection as shown in Fig.~\ref{fig:RQA-exps}. During execution, the VLM is queried sequentially to provide guidance at each stage, and the human operator proceeds based on the model’s responses unless human intervention is required.

\noindent\textbf{Results}.
We evaluate multiple vision-language models by asking the designed questions in sequence and measuring how far the knot-tying task can progress before intervention is necessary. Results are summarized in Table~\ref{tab:comparison}. We quantify execution progress with at most 3 human interventions. Across all models, higher RoboVista-surgical scores are associated with greater shared autonomy task progress. Models with stronger benchmark performance consistently complete more stages of the knot-tying procedure before requiring assistance.
All models exhibit a common failure mode at the second question that requires pointing, where motion execution fails despite correct high-level reasoning. The visualization is illustrated in the second image of the first row in Fig.~\ref{fig:RQA-exps}. The failure suggests that even minor execution errors can be amplified across consecutive decisions and the overall physical task progress is correlated with RoboVista-surgical benchmark performance.

\begin{table}[t]
\centering
\scriptsize
\setlength{\tabcolsep}{3.5pt}     
\caption{\textbf{Surgical knot-tying performance under shared autonomy.}
\emph{RoboVista Surgery} denotes benchmark accuracy on surgical Robot-VQA tasks.
\emph{Progress w/o Intervention} is the furthest knot-tying stage completed without intervention),
and \emph{Progress w/Intervention} is the furthest stage completed with up to three interventions.}
\begin{tabular}{lccc}
\toprule
\textbf{Model} &
\makecell{\textbf{RoboVista-Surgery}\\\textbf{}} &
\makecell{\textbf{Progress}\\\textbf{w/o Intervention}} &
\makecell{\textbf{Progress}\\\textbf{w/ Intervention}} \\
\midrule
Qwen-2.5 32B    & 58.7 & 2/16  & 9/16  \\
ChatGPT-5.0     & 63.0 & 2/16  & 9/16  \\
Gemini 2.5 Pro  & \textbf{76.1} & 2/16  & \textbf{15/16} \\
\bottomrule
\end{tabular}

\label{tab:comparison}
\end{table}

%% file: chapters/conclusion.tex
\section{Conclusion}

We presented {Robot Question Answering (RQA)}, a modular framework that translates diverse robot application decision points into a unified, robot-centric VQA interface, and instantiated it as {RoboVista}, a curated benchmark spanning six real-world robotic domains and 39 task types. Through comprehensive evaluation, we show that state-of-the-art VLMs exhibit substantial and persistent performance gaps across domains and reasoning stages. 

\textbf{Limitations.} Constructing high-quality Robot-VQA instances requires significant effort from human domain experts to ensure visual grounding, task relevance, and unambiguous answers. While parts of the pipeline can be automated, in future work we will explore automated question generation and verification, and extend \benchname to additional robot tasks, embodiments, and interactions.

%% file: chapters/appendix.tex
\subsection{Key Dataset Statistics}
Table~\ref{tab:dataset-stats} summarizes the key statistics of  \benchname dataset. The dataset comprises 474 samples with 730 total images, averaging 1.54 images per question across 33 unique tasks from 62 publication sources. The samples span six       
  robotic domains: open datasets (31.6\%), industrial manufacturing (30.4\%), agriculture (13.1\%), domestic robotics (11.0\%), surgical robotics (9.7\%), and autonomous driving (4.2\%). The majority of questions assess scene understanding abilities (66.9\%), 
  followed by high-level decision making (16.7\%) and low-level motion awareness (14.3\%). The correct answer distribution across choices A through E is relatively balanced (24\%, 25\%, 22\%, 18\%, 11\%), mitigating potential biases in answer selection. Most  
  samples contain a single image (74.9\%), while 25.1\% feature multiple images for tasks requiring temporal or multi-view reasoning. Questions average 25.1 words in length, with expert-provided reasoning averaging 41.9 words per sample. 
\input{tables/table_overview}

\subsection{Key Inference Parameters}
\paragraph{Evaluation Infrastructure.}
We deploy all open-source vision-language models using SGLang equipped with eight NVIDIA A100 GPUs (80GB each). For proprietary models (GPT-4o, GPT-5, Gemini-2.5-Pro), we use their respective official APIs. All experiments are conducted with deterministic decoding (temperature 0.0, top-p 1.0) and a maximum output length of 10,240 tokens to ensure reproducibility.

To ensure consistent input across models with varying native resolutions, we preprocess all images by resizing them such that the maximum dimension (width or height) does not exceed 720 pixels while preserving the original aspect ratio. Images are then converted to JPEG format with quality 85 and encoded as base64 strings for API transmission. For questions with multiple images, each image is processed independently and provided to the model in sequence.

\begin{table}
\centering
\caption{Evaluation Parameters}
\label{tab:eval-params}
\begin{tabular}{ll}
\toprule
\textbf{Parameter} & \textbf{Value} \\
\midrule
\multicolumn{2}{l}{\textit{Hardware \& Serving}} \\
Accelerator & NVIDIA A100 (80GB) $\times$ 8 \\
Cloud Platform & Google Cloud Platform \\
\midrule
\multicolumn{2}{l}{\textit{Generation Settings}} \\
Max Context Length & 32,768 tokens \\
Max Output Tokens & 10,240 tokens \\
Temperature & 0.0 (greedy decoding) \\
Top-p & 1.0 \\
\midrule
\multicolumn{2}{l}{\textit{Image Processing}} \\
Max Image Resolution & 720 px \\
Image Format & JPEG (quality 85) \\
\bottomrule
\end{tabular}
\end{table}

\textbf{Prompt Design.}
Each question is formatted with the question text followed by five multiple-choice options labeled A through E, presented as:
\begin{verbatim}
{question}

Options:
A. {choice_a}
B. {choice_b}
C. {choice_c}
D. {choice_d}
E. {choice_e}

{suffix}
\end{verbatim}
We evaluate models under three prompting strategies. 

For \textit{Standard} (zero-shot) prompting, the system prompt instructs: ``You are a helpful assistant analyzing robotic manipulation images. Answer the multiple choice question based on the images. Respond with only the letter of your answer,'' with the suffix ``Answer with the letter only (A, B, C, D, or E).'' 

For \textit{Chain-of-Thought (CoT)} prompting, the system prompt omits the single-letter constraint, and the suffix instructs: ``Think step by step about this question. Put your reasoning inside \texttt{<think>}...\texttt{</think>} tags, then provide your final answer as a single letter (A, B, C, D, or E).'' For Qwen3 models with native reasoning capabilities, we explicitly disable the built-in thinking mode via \texttt{chat\_template\_kwargs} to ensure fair comparison.

For \textit{In-Context Learning (ICL).}
To investigate whether models can benefit from task-specific demonstrations, we evaluate two in-context learning configurations. For each test question, we retrieve an example from the same task and domain by building an index of questions grouped by (task, domain) pairs and randomly sampling one example (excluding the test question itself) using a fixed seed for reproducibility. The prompt is structured as:
\begin{verbatim}
=== EXAMPLE ===
[Image 1: Example image]

Question: {example_question}
Options: {example_choices}
Answer: {correct_answer}

=== YOUR QUESTION ===
[Image 2: Your image]
{test_question}
\end{verbatim}
We test {ICL without reasoning}, where only the correct answer is provided without explanation. The system prompt for ICL conditions states: ``You will be given an example question with its answer, followed by a new question to answer. Use the example to help you understand the task.'' For ICL experiments, we additionally request confidence scores in the format ``Confidence: 0-100\%'' to analyze model calibration. To account for variance in example selection, we support running multiple trials with different random seeds and report averaged results with standard deviation.

\paragraph{Answer Extraction.}
We employ a robust multi-stage regex-based parser to extract the predicted answer from model responses. The parser first checks for content following \texttt{</think>} tags to handle CoT responses. It then applies a sequence of patterns to identify the answer: (1) explicit statements such as ``the answer is X'' or ``X is the correct answer''; (2) standalone letters at the beginning or end of the response; (3) letters followed by periods or parentheses; and (4) as a fallback, any isolated letter A--E appearing in the response. For ICL experiments, we additionally parse confidence scores using patterns like ``Confidence: N\%'' and normalize values to the [0, 1] range. This hierarchical approach ensures reliable extraction across diverse response formats.

\subsection{Additional Results}

\begin{figure}
    \centering
    \includegraphics[width=\linewidth]{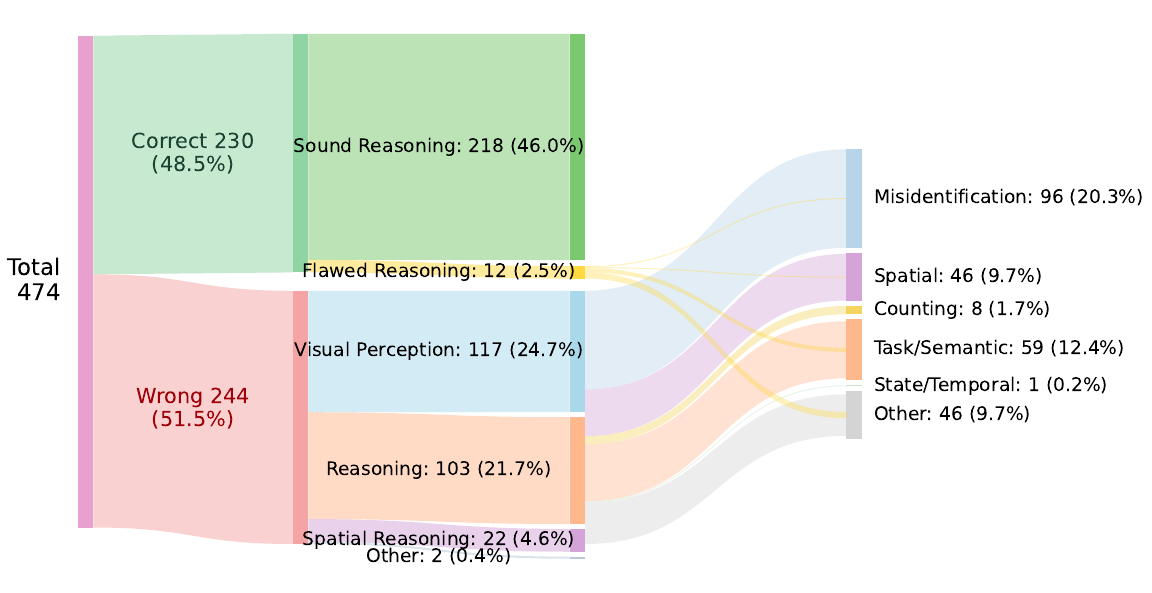}
    \caption{\textbf{Failure Analysis of Qwen2.5-VL 7B}. We analyze the reasoning chain of Qwen2.5-VL 7B and categorize failures in misidentification and reasoning.  }
    \label{fig:failure}
\end{figure}

\begin{figure}
    \centering
    \includegraphics[width=\linewidth]{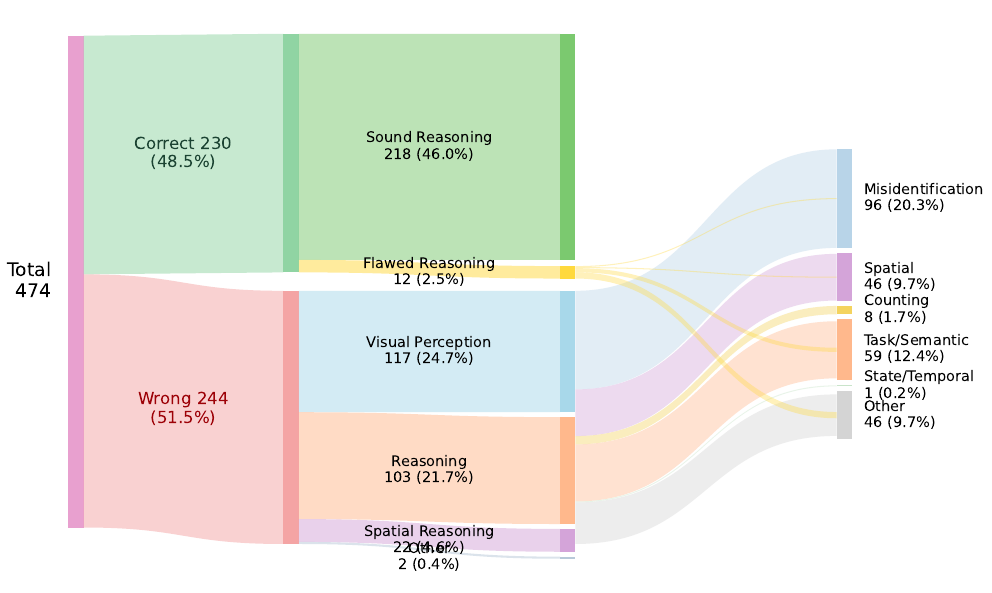}
    \caption{\textbf{Failure Analysis of Qwen3-VL 235B}. We analyze the reasoning chain of Qwen3-VL 235B and categorize failures in misidentification and reasoning.  }
    \label{fig:placeholder}
\end{figure}

\textbf{Results on Failure Analysis}
We conduct failure analysis using model-generated reasoning chains that separates visual perception errors from higher-level reasoning failures.  For both correct and incorrect predictions, we use Qwen3-VL-32B-Thinking to propose an initial failure category given groundtruth human reasoning trace and answer. Then we manually verify and correct these assignments.

Figure~\ref{fig:failure} shows the failure breakdown. 
A key observation is that the majority of failures originate from visual perception rather than logical reasoning.
The most frequent failure mode is misidentification, accounting for 143 cases (30.2\%), primarily involving incorrect object identity, state, or spatial location.
Common examples include confusing visually similar objects, or incorrectly localizing target objects under occlusion or clutter.
Spatial reasoning errors form the second largest category, encompassing failures in relative positioning, depth estimation, motion inference, and reachability.
Task and semantic misinterpretation reflects failures in mapping visual evidence to task intent, such as misunderstanding which object is being manipulated or misinterpreting contact and interaction states.

We perform the same analysis for the much larger 235B model.
Scaling significantly reduces both the overall error rate and the proportion of flawed reasoning chains, with correct answers increasing from 36.5\% to 48.5\%.
Misidentification errors drop substantially (from 30.2\% to 20.3\%), indicating improved visual recognition and grounding at scale.
However, spatial and semantic errors persist, and remain a dominant source of failure even for the largest model.
This suggests that while increased model capacity improves visual robustness and reasoning consistency, fundamental challenges in spatial understanding and geometry-aware perception are not fully resolved by scale alone.

\input{tables/cot_comprehensive}

\textbf{CoT Prompting by Domain}
The table reports the accuracy difference between Chain-of-Thought (CoT) prompting and zero-shot prompting across a wide range of vision–language models, broken down by task domain, and separately for scene understanding (top) and planning (bottom). Overall, CoT exhibits a clear asymmetry: for scene understanding, CoT often degrades performance, especially for smaller and mid-sized models, with consistent negative deltas across agriculture, home, and open datasets, indicating that explicit reasoning may distract from perceptual grounding. Gains are sparse and domain-specific, appearing mainly in driving and industry for larger or more recent models. In contrast, for planning, CoT is substantially more beneficial, with many models showing positive improvements in aggregate and pronounced gains in structured domains such as agriculture, driving, and surgery. Larger models tend to benefit more consistently from CoT in planning, while smaller models show mixed or unstable behavior. Taken together, the results suggest that CoT is not universally helpful in embodied settings: it is often detrimental for low-level scene understanding, but can meaningfully improve higher-level planning when the task requires explicit reasoning over actions, constraints, and temporal structure.

\subsection{Physical Evaluation: Distance and Positional Estimation}
\begin{figure}
    \centering
    \includegraphics[width=\linewidth]{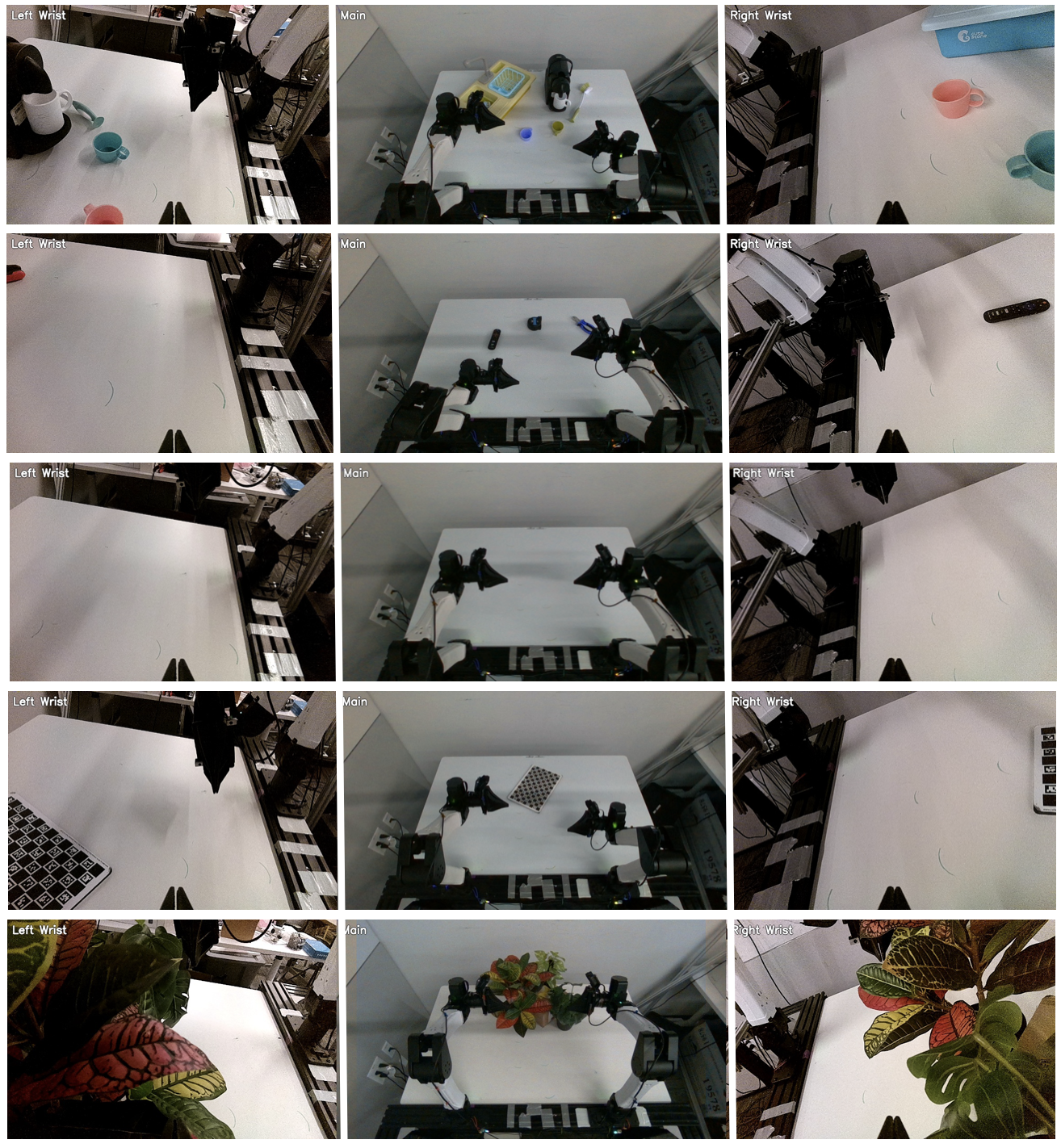}
    \caption{\textbf{Evaluation Setup of Distance Estimation}. From top to bottom is kitchen, industrial, empty, checkerboard, and plants.}
    \label{fig:bimanual:5}
\end{figure}

As shown in Fig.~\ref{fig:bimanual:5}, we evaluate VLMs on a bimanual manipulation task that requires reasoning about spatial relationships and generating symbolic motion plans. The visual input consists of three synchronized camera views: a left wrist-mounted camera, a right wrist-mounted camera, and a top-down main camera providing an overhead perspective of the workspace, concatenated into a single composite image. Given these observations, the VLM is tasked with two objectives: (1) estimating the Euclidean distance between the left and right gripper tips in centimeters, and (2) generating a sequence of axis-aligned symbolic motion commands that would move the right gripper to align tip-to-tip with the left gripper.

The language prompt defines a robot-centric coordinate system relative to the main camera view: forward/backward correspond to movement toward/away from the top/bottom of the image ($x$-axis), left/right correspond to lateral movement ($y$-axis), and up/down correspond to depth movement toward/away from the camera ($z$-axis). The model must output its response in a strict format:
\begin{verbatim}
DISTANCE: <value> cm
COMMAND: [("<direction>", <amount>, "cm")]
\end{verbatim}
where each command tuple specifies a direction (forward, backward, left, right, up, down) and a metric distance. The prompt explicitly instructs models to leverage visible background objects as geometric priors for scale estimation, rather than relying on arbitrary placeholder values.

We evaluate across five distinct scene configurations to test generalization across visual contexts: (1) \textit{Kitchen Objects}---common household items including cups and containers; (2) \textit{Small Industrial Parts}---small manipulable objects such as remote controls; (3) \textit{No Background Objects}---an empty white table surface; (4) \textit{Checkerboard}---a calibration pattern providing regular geometric cues; and (5) \textit{Plants}---organic, irregularly-shaped objects. For each scene configuration, we collect 5 samples with the left and right arms randomly positioned to span the reachable workspace, yielding inter-gripper distances ranging from approximately 10 to 40\,cm. Ground-truth tip positions are computed using the robot's joint encoder readings and forward kinematics of the ALOHA bimanual platform. We report three metrics: (1) \textit{distance estimation error}---the signed difference between the VLM's estimated distance and the ground-truth Euclidean distance; (2) \textit{command magnitude}---the total distance specified by the VLM's motion commands, computed as $\sqrt{\Delta x^2 + \Delta y^2 + \Delta z^2}$; and (3) \textit{position error}---the Euclidean distance between the predicted right gripper position (after applying the VLM's commands to the actual right tip position) and the target left gripper position. To account for stochasticity in model responses, we execute 5 independent runs per sample and report averaged metrics with standard deviation.

\subsection{Physical Evaluation: Surgical Knot-Tying using dVRK}

To evaluate the relationship between benchmark performance and real-world task success in a safety-critical domain, we conduct closed-loop surgical knot-tying experiments using the da Vinci Research Kit (dVRK). This benchmark assesses whether performance on RoboVQA correlates with a model's ability to support long-horizon, physically grounded task execution in surgical scenarios.

We consider a shared-autonomy setting motivated by scenarios where a VLM assists junior operators during surgical task execution. The physical setup uses an emulated environment to allow controlled evaluation while preserving the visual and procedural structure of real knot-tying workflows. The experimental scene consists of two surgical grippers (Patient Side Manipulators), a curved surgical needle, suture thread, and a wound phantom representing tissue to be sutured. Visual input is provided via the dVRK's stereo endoscope, with left and right camera streams published as ROS topics at the native resolution.

Human experts with surgical robotics experience designed a sequence of 16 VQA questions corresponding to critical decision points throughout the knot-tying procedure, as shown in Table~\ref{tab:dvrk-questions}.

\begin{table}[H]
\centering
\small
\caption{Surgical knot-tying VQA questions for dVRK shared-autonomy evaluation.}
\label{tab:dvrk-questions}
\resizebox{\columnwidth}{!}{%
\begin{tabular}{clll}
\toprule
\textbf{Q} & \textbf{Question} & \textbf{Type} & \textbf{Answer} \\
\midrule
1 & Is needle held securely by right gripper? & Binary & Yes \\
2 & Annotate optimal grasping point on needle & Spatial & Coordinates \\
3 & Has left gripper clamped the needle body? & Binary & Yes \\
4 & Pull needle left or right for extraction? & Direction & Left \\
5 & Has right gripper clamped the thread? & Binary & Yes \\
6 & Position left gripper inside/above/below loop? & Choice & Inside \\
7 & Is medial or lateral gripper orientation better? & Choice & Medial \\
8 & Wrap clockwise or counter-clockwise? & Direction & Clockwise \\
9 & Has loop been established around left gripper? & Binary & Yes \\
10 & Annotate grasping point on suture tail & Spatial & Coordinates \\
11 & Move left gripper left or right to capture tail? & Direction & Right \\
12 & Has left gripper clamped the suture tail? & Binary & Yes \\
13 & Move left gripper left or right to tighten? & Direction & Left \\
14 & Move right gripper left or right to tighten? & Direction & Right \\
15 & Is suture over-pulled, under-pulled, or proper? & Choice & Over-pulled \\
16 & Annotate two cutting points on suture tails & Spatial & Coordinates \\
\bottomrule
\end{tabular}
}
\end{table}

During execution, the VLM is integrated via a ROS node that subscribes to the dVRK stereo endoscope camera topics and publishes structured inference results. Images are resized to a maximum of 1024 pixels on the longest dimension and encoded as JPEG (quality 85) before transmission to the VLM API.
\begin{figure}
    \centering
    \includegraphics[width=\linewidth]{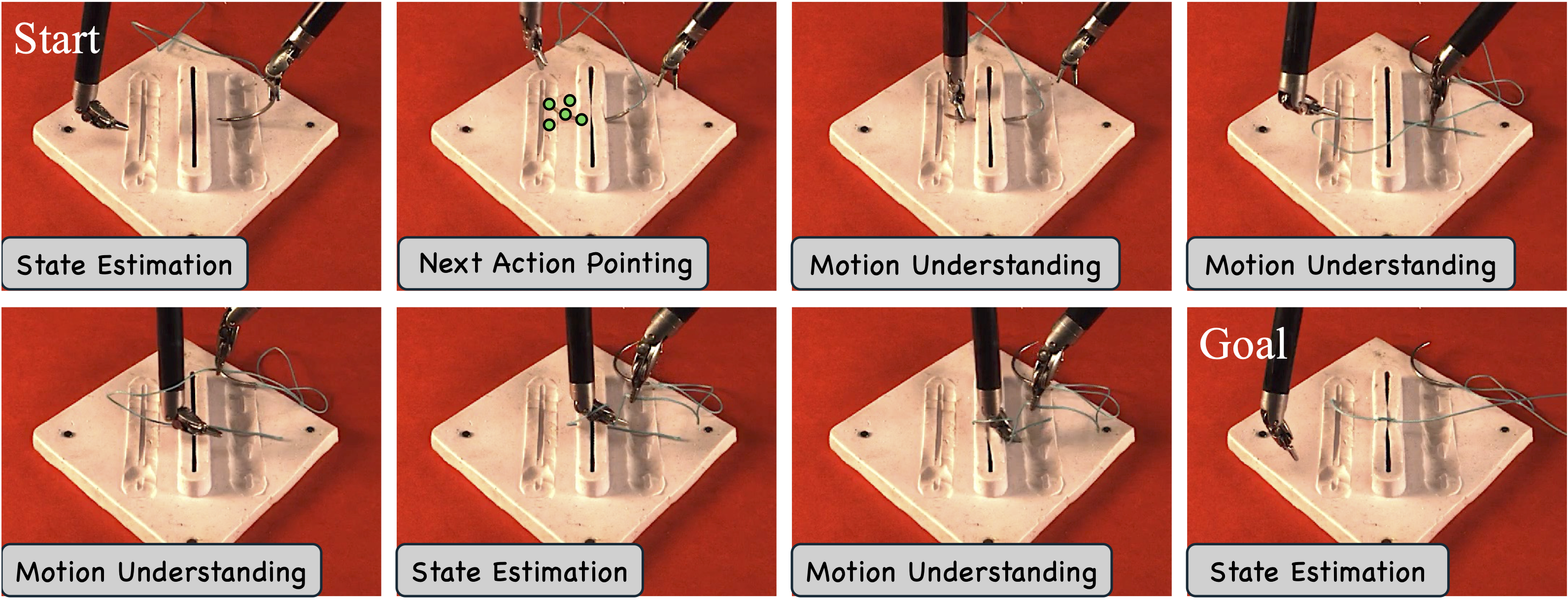}
    \caption{\textbf{Visual Questions for surgical knot tying}. }
    \label{fig:knot}
\end{figure}

For spatial annotation questions (Q2, Q10, Q16), the prompt is augmented with explicit instructions to output pixel coordinates in JSON format: \texttt{\{``x'': <pixel\_x>, ``y'': <pixel\_y>\}} along with the image dimensions. A multi-strategy coordinate parser extracts points from the VLM response, supporting both single coordinates and arrays with labels. Parsed coordinates are scaled back to the original image resolution and visualized as colored crosshair markers overlaid on the camera feed.

The system prompt establishes task context: ``You are an expert surgical assistant AI monitoring an emulated robotic surgery task on the da Vinci Research Kit robot. The experimental scene contains two grippers, a surgical needle, a suture thread, and a wound phantom. Your task is to monitor and guide a teleoperated surgical knot tying procedure.'' The VLM is queried sequentially at each procedural stage, and the human operator proceeds based on the model's responses unless intervention is required due to incorrect guidance. We record: (1) \textit{task completion rate}---whether the knot-tying procedure was successfully completed; (2) \textit{VLM accuracy}---the fraction of queries answered correctly without human override; (3) \textit{intervention count}---the number of times the operator had to disregard VLM guidance; and (4) \textit{per-question accuracy}---correctness rates stratified by question type (binary, directional, spatial).

\subsection{Visual Question Answering Examples}

The following figures illustrate several VQA tasks conducted using robotic trajectories. Each figure presents a unique scenario where human expertise was used to validate the correctness of robotic actions or spatial understanding based on visual inspection.

\vqafigurefixed{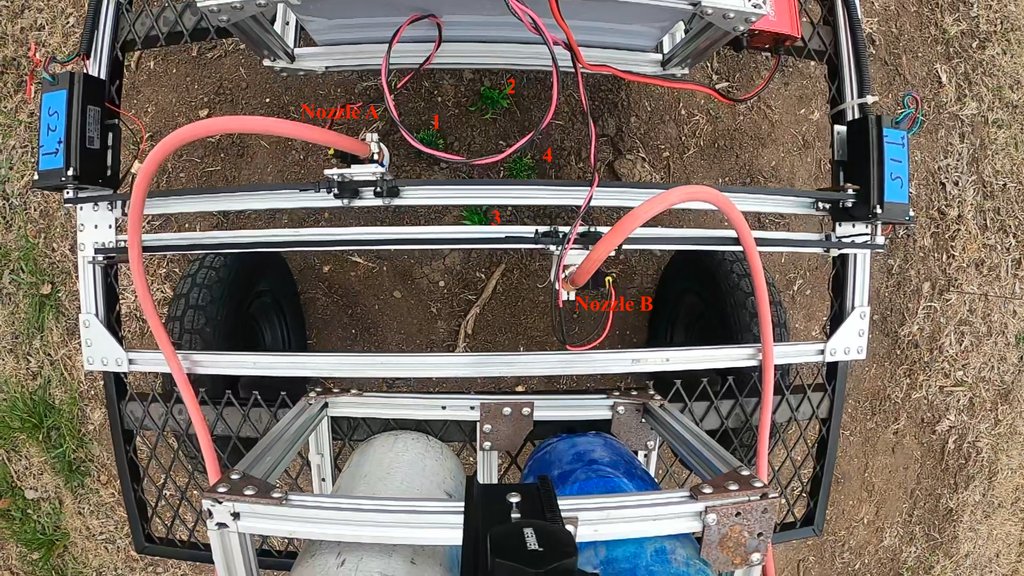}%
{The image shows a weed-removal robot equipped with two independently controlled spray nozzles, Nozzle A and Nozzle B, mounted on a horizontal rail. Each nozzle can move left/right (horizontally in the image). The robot platform moves forward along the vertical direction of the image (from bottom to top). There are four weeds, labeled 1 through 4. Task: Select the following choice that assigns the two nozzles spray all four weeds most efficiently. \\\emph{(A) Nozzle A: 1, 3 Nozzle B: 2, 4 (B) Nozzle A: 1, 4 Nozzle B: 2, 3 (C) Nozzle A: 2, 3 Nozzle B: 1, 4 (D) Nozzle A: 2, 4 Nozzle B: 1, 3 (E) Nozzle A: 1, 2 Nozzle B: 3, 4}}%
{(E) Nozzle A: 1, 2 Nozzle B: 3, 4}%
{Efficiency is defined by three criteria: each weed is sprayed exactly once; each nozzle handles nearby weeds aligned with the robot’s forward motion to reduce switching and lateral movement; and weeds are sprayed in the order encountered. Under these constraints, only option E satisfies all requirements.}%
{VQA example: Weed Removal (high level decision making, goal condition action reasoning)}

\include{latex_examples/vqa_examples}

%% file: tables/table_overview.tex
\begin{table}[H]
\centering
\caption{\benchname Dataset Statistics}
\label{tab:dataset-stats}
\resizebox{\columnwidth}{!}{%
\begin{tabular}{lrl}
\toprule
\textbf{Item} & \textbf{Count} & \textbf{\%} \\
\midrule
Total Samples & 474 & -- \\
Total Images & 730 & -- \\
Avg Images/Sample & 1.54 & -- \\
Unique Tasks & 33 & -- \\
Unique Sources & 62 & -- \\
\midrule
open datasets & 150 & 31.6\% \\
industrial manufacturing & 144 & 30.4\% \\
agriculture & 62 & 13.1\% \\
domestic & 52 & 11.0\% \\
surgical robotics & 46 & 9.7\% \\
autonomous driving & 20 & 4.2\% \\
\midrule
scene understanding & 317 & 66.9\% \\
high level decision making & 79 & 16.7\% \\
low level motion awareness & 68 & 14.3\% \\
recovery replanning robustness & 9 & 1.9\% \\
physical interaction & 1 & 0.2\% \\
\midrule
Correct Answer (A/B/C/D/E) & 114/117/105/85/53 & 24/25/22/18/11\% \\
\midrule
1 image & 355 & 74.9\% \\
2 images & 66 & 13.9\% \\
3+ images & 53 & 11.2\% \\
\midrule
Question length (avg words) & 25.1 & -- \\
Reasoning length (avg words) & 41.9 & -- \\
\bottomrule
\end{tabular}%
}
\end{table}

%% file: tables/cot_comprehensive.tex

\begin{table}[t]
    \centering

    \captionsetup{font=footnotesize}
    \setlength{\tabcolsep}{4pt}

    \begin{subtable}{\linewidth}
        \centering
        \caption{\textbf{Does Chain-of-Thought Improve Scene Understanding?} Showing accuracy difference (CoT $-$ Zero-Shot). \textcolor{ForestGreen}{Green} = improvement, \textcolor{red}{Red} = degradation.}
        \resizebox{\linewidth}{!}{%
        \begin{tabular}{@{}l c c c c c c c @{}}
        \toprule
        \textbf{Model} & \textbf{All} & \cellcolor{green!15}\textbf{Agriculture} & \cellcolor{blue!15}\textbf{Driving} & \cellcolor{yellow!15}\textbf{Home} & \cellcolor{gray!15}\textbf{Industry} & \cellcolor{red!15}\textbf{Surgery} & \cellcolor{orange!15}\textbf{Open Datasets} \\
        \midrule
        Qwen 2.5 VL-3B & \textcolor{red}{-11.9} & \textcolor{red}{-6.1} & \textcolor{red}{-33.3} & \textcolor{red}{-25.5} & \textcolor{red}{-6.7} & \textcolor{red}{-20.0} & \textcolor{red}{-7.7} \\
        Qwen 2.5 VL-7B & \textcolor{red}{-11.3} & \textcolor{red}{-10.2} & \textcolor{red}{-11.1} & \textcolor{red}{-13.7} & \textcolor{red}{-12.0} & \textcolor{red}{-13.3} & \textcolor{red}{-9.6} \\
        Qwen 2.5 VL-32B & \textcolor{red}{-6.3} & \textcolor{red}{-14.3} & \textcolor{red}{-11.1} & \textcolor{red}{-11.8} & \textcolor{red}{-2.7} & \textcolor{ForestGreen}{+3.3} & \textcolor{red}{-4.8} \\
        Qwen 2.5 VL-72B & \textcolor{red}{-4.4} & \textcolor{red}{-12.2} & \textcolor{ForestGreen}{+11.1} & \textcolor{red}{-3.9} & \textcolor{ForestGreen}{+1.3} & +0.0 & \textcolor{red}{-7.7} \\
        Robo2VLM-ER & \textcolor{red}{-8.8} & \textcolor{ForestGreen}{+4.1} & \textcolor{ForestGreen}{+22.2} & \textcolor{red}{-9.8} & \textcolor{red}{-14.7} & \textcolor{red}{-26.7} & \textcolor{red}{-7.7} \\
        RoboBrain 2.5-8B & \textcolor{red}{-6.9} & \textcolor{red}{-10.2} & +0.0 & \textcolor{red}{-11.8} & \textcolor{red}{-8.0} & +0.0 & \textcolor{red}{-4.8} \\
        RoboBrain 2.5-8B-NV & \textcolor{red}{-7.2} & \textcolor{ForestGreen}{+12.2} & \textcolor{red}{-11.1} & \textcolor{red}{-5.9} & \textcolor{red}{-17.3} & \textcolor{red}{-20.0} & \textcolor{red}{-5.8} \\
        Qwen 3 VL-4B & \textcolor{red}{-1.3} & \textcolor{ForestGreen}{+6.1} & \textcolor{ForestGreen}{+22.2} & \textcolor{red}{-9.8} & \textcolor{red}{-2.7} & \textcolor{ForestGreen}{+13.3} & \textcolor{red}{-5.8} \\
        Qwen 3 VL-8B & \textcolor{red}{-0.3} & \textcolor{ForestGreen}{+4.1} & \textcolor{ForestGreen}{+33.3} & \textcolor{red}{-13.7} & \textcolor{ForestGreen}{+2.7} & \textcolor{red}{-6.7} & \textcolor{ForestGreen}{+1.0} \\
        Qwen 3 VL-8B-T & \textcolor{ForestGreen}{+2.2} & \textcolor{red}{-8.2} & +0.0 & \textcolor{ForestGreen}{+3.9} & \textcolor{ForestGreen}{+6.7} & +0.0 & \textcolor{ForestGreen}{+3.8} \\
        Qwen 3 VL-32B-T & \textcolor{red}{-1.6} & \textcolor{red}{-4.1} & \textcolor{ForestGreen}{+22.2} & \textcolor{red}{-5.9} & \textcolor{red}{-1.3} & \textcolor{red}{-10.0} & \textcolor{ForestGreen}{+1.9} \\
        Qwen 3-235B-A22B & \textcolor{red}{-2.5} & \textcolor{red}{-2.0} & \textcolor{ForestGreen}{+11.1} & \textcolor{red}{-2.0} & \textcolor{red}{-1.3} & \textcolor{red}{-6.7} & \textcolor{red}{-3.8} \\
        GPT-5 & \textcolor{red}{-0.3} & \textcolor{red}{-12.2} & \textcolor{red}{-11.1} & \textcolor{red}{-2.0} & \textcolor{ForestGreen}{+8.0} & +0.0 & \textcolor{ForestGreen}{+1.0} \\
        \bottomrule
        \end{tabular}
        }%
        \label{tab:cot-sc}
    \end{subtable}

    \vspace{0.6em}

    \begin{subtable}{\linewidth}
        \centering
        \caption{\textbf{Does Chain-of-Thought Improve Planning?} Showing accuracy difference (CoT $-$ Zero-Shot). \textcolor{ForestGreen}{Green} = improvement, \textcolor{red}{Red} = degradation.}
        \resizebox{\linewidth}{!}{%
        \begin{tabular}{@{}l c c c c c c c @{}}
        \toprule
        \textbf{Model} & \textbf{All} & \cellcolor{green!15}\textbf{Agriculture} & \cellcolor{blue!15}\textbf{Driving} & \cellcolor{yellow!15}\textbf{Home} & \cellcolor{gray!15}\textbf{Industry} & \cellcolor{red!15}\textbf{Surgery} & \cellcolor{orange!15}\textbf{Open Datasets} \\
        \midrule
        Qwen 2.5 VL-3B & \textcolor{red}{-7.1} & \textcolor{red}{-15.4} & \textcolor{ForestGreen}{+18.2} & +0.0 & \textcolor{red}{-3.9} & \textcolor{ForestGreen}{+18.8} & \textcolor{red}{-30.0} \\
        Qwen 2.5 VL-7B & \textcolor{ForestGreen}{+1.3} & \textcolor{red}{-7.7} & \textcolor{red}{-9.1} & \textcolor{ForestGreen}{+8.0} & \textcolor{ForestGreen}{+2.0} & \textcolor{ForestGreen}{+18.8} & \textcolor{red}{-5.0} \\
        Qwen 2.5 VL-32B & \textcolor{red}{-6.4} & \textcolor{red}{-23.1} & \textcolor{red}{-9.1} & \textcolor{red}{-12.0} & \textcolor{red}{-3.9} & \textcolor{ForestGreen}{+6.2} & \textcolor{red}{-5.0} \\
        Qwen 2.5 VL-72B & \textcolor{ForestGreen}{+5.1} & \textcolor{ForestGreen}{+15.4} & +0.0 & \textcolor{ForestGreen}{+8.0} & \textcolor{ForestGreen}{+5.9} & +0.0 & \textcolor{ForestGreen}{+2.5} \\
        Robo2VLM-ER & \textcolor{ForestGreen}{+2.6} & \textcolor{ForestGreen}{+7.7} & +0.0 & \textcolor{red}{-16.0} & \textcolor{ForestGreen}{+9.8} & \textcolor{ForestGreen}{+12.5} & +0.0 \\
        RoboBrain 2.5-8B & \textcolor{red}{-1.3} & \textcolor{ForestGreen}{+30.8} & \textcolor{red}{-9.1} & \textcolor{red}{-4.0} & \textcolor{red}{-5.9} & \textcolor{ForestGreen}{+18.8} & \textcolor{red}{-10.0} \\
        RoboBrain 2.5-8B-NV & \textcolor{red}{-1.9} & +0.0 & \textcolor{ForestGreen}{+9.1} & \textcolor{red}{-8.0} & +0.0 & \textcolor{red}{-6.2} & \textcolor{red}{-2.5} \\
        Qwen 3 VL-4B & \textcolor{ForestGreen}{+4.5} & \textcolor{ForestGreen}{+7.7} & +0.0 & \textcolor{ForestGreen}{+8.0} & \textcolor{red}{-9.8} & \textcolor{ForestGreen}{+25.0} & \textcolor{ForestGreen}{+12.5} \\
        Qwen 3 VL-8B & \textcolor{red}{-3.2} & \textcolor{red}{-23.1} & +0.0 & +0.0 & \textcolor{red}{-3.9} & +0.0 & +0.0 \\
        Qwen 3 VL-32B & \textcolor{ForestGreen}{+8.3} & \textcolor{ForestGreen}{+7.7} & \textcolor{ForestGreen}{+9.1} & +0.0 & \textcolor{ForestGreen}{+11.8} & +0.0 & \textcolor{ForestGreen}{+12.5} \\
        Qwen 3-235B-A22B & \textcolor{red}{-2.6} & \textcolor{ForestGreen}{+15.4} & \textcolor{red}{-18.2} & \textcolor{ForestGreen}{+4.0} & \textcolor{red}{-2.0} & \textcolor{red}{-6.2} & \textcolor{red}{-7.5} \\
        GPT-5 & \textcolor{ForestGreen}{+1.3} & +0.0 & \textcolor{ForestGreen}{+18.2} & \textcolor{red}{-12.0} & \textcolor{ForestGreen}{+2.0} & \textcolor{ForestGreen}{+12.5} & +0.0 \\
        \bottomrule
        \end{tabular}
        }%
        \label{tab:cot-pl}
    \end{subtable}

\end{table}

%% file: latex_examples/vqa_examples.tex




\vqafigure{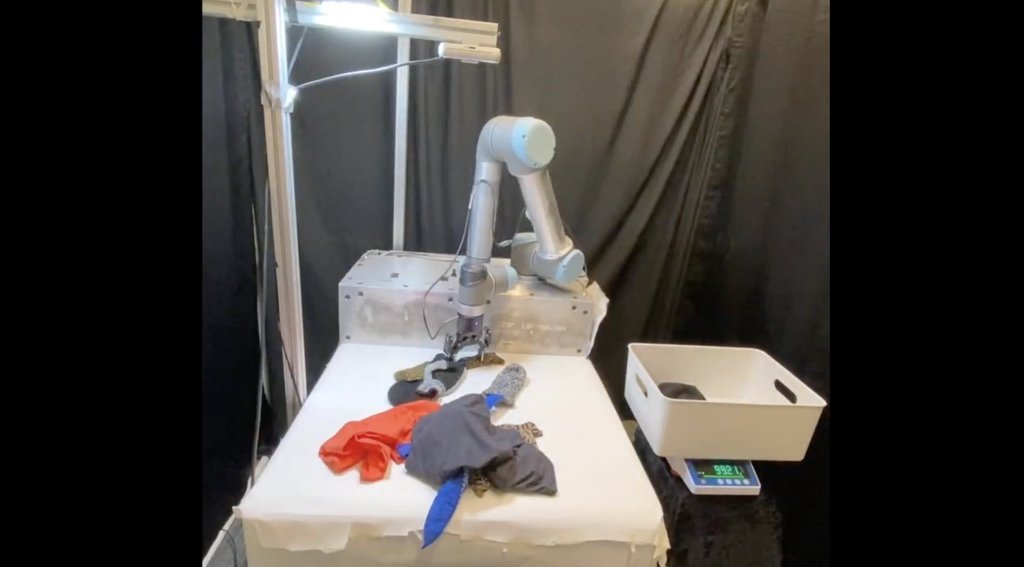}%
{The robot is trying to maximize its picking efficiency by grasping multiple garments at the same time. Analyze the workspace and the grasp it is currently attempting. Which of the following is the most correct? \\\emph{(A) The robot will only pick up the gray glove and the yellow/ green sock near its gripper. (B) The robot will pick up the large bright red shirt located at the front left of the pile. (C) The robot will only pick up the yellow/ green sock near its gripper. (D) The robot will pick up the gray glove, the yellow/ green sock near its gripper, and the black sock under it. (E) The robot has likely failed this grasp attempt because the gripper jaws are already fully closed and hovering above the clothes.}}%
{(A) The robot will only pick up the gray glove and the yellow/ green sock near its gripper.}%
{The robot's grippers are pushed against the glove and the yellow/ green sock. The black sock is too far away and cannot be picked up.}%
{VQA example: Garmet Manipulation (high level decision making, goal condition action reasoning)}


\vqafigure{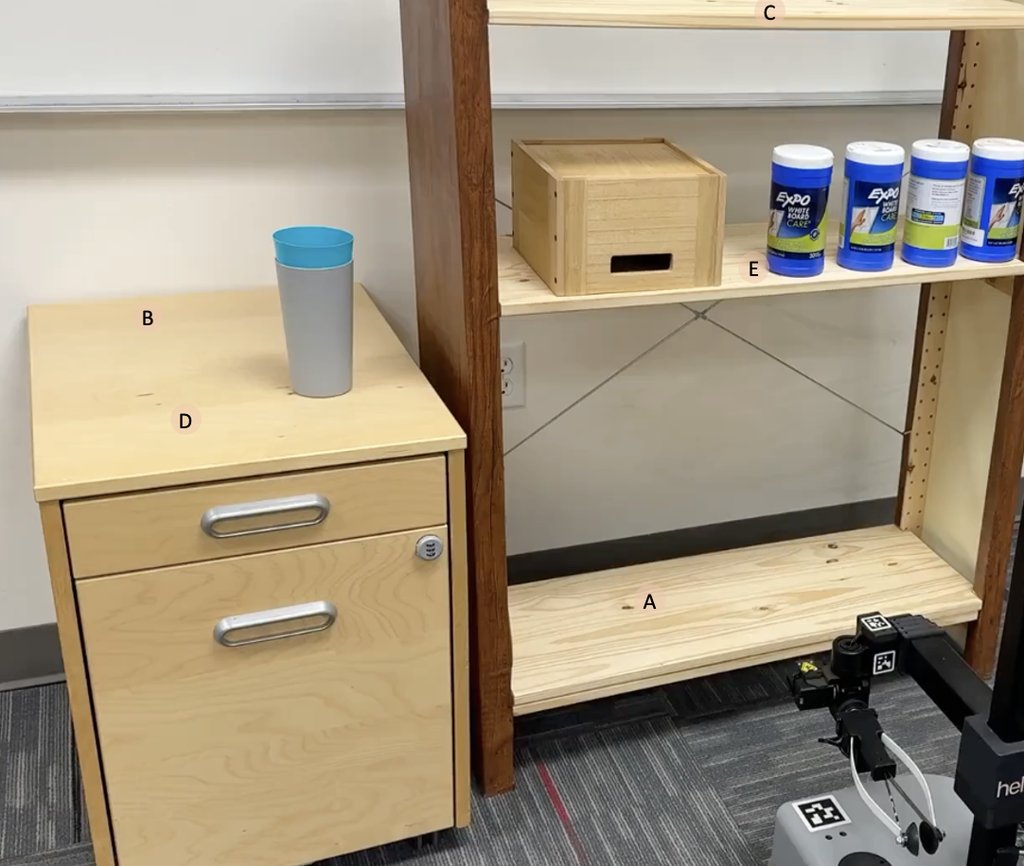}%
{I want to place the stack of cups into the middle layer of the shelf. Could you point me to the middle layer of the shelf with several wipers? \\\emph{(A) A (B) B (C) C (D) D (E) E}}%
{(E) E}%
{instance segmentation. Semantic segmentation. Spatial reasoning.}%
{VQA example: Tidybot (scene understanding, geometry spatial reasoning)}

\vqafigure{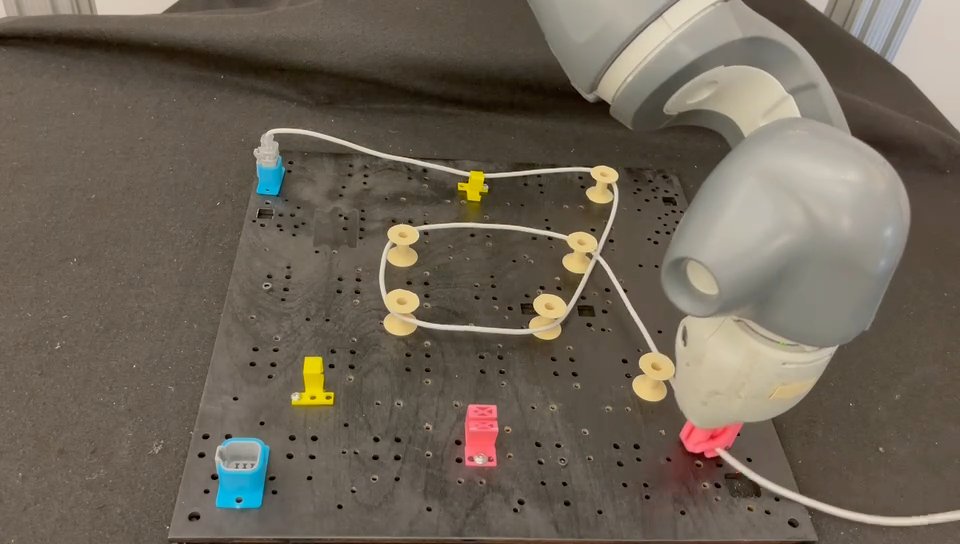}%
{To insert the cable into the pink fixture, what is the required movement of the gripper? \\\emph{(A) Move right (B) Move Left and Down (C) Move upward (D) Move down (E) Stay Stationary}}%
{(B) Move Left and Down}%
{The pink fixture is located to the left of the gripper’s current position and requires downward insertion into the clip aperture.}%
{VQA example: Cable Routing (low level motion awareness, motion feasibility collision avoidance)}

\vqafigure{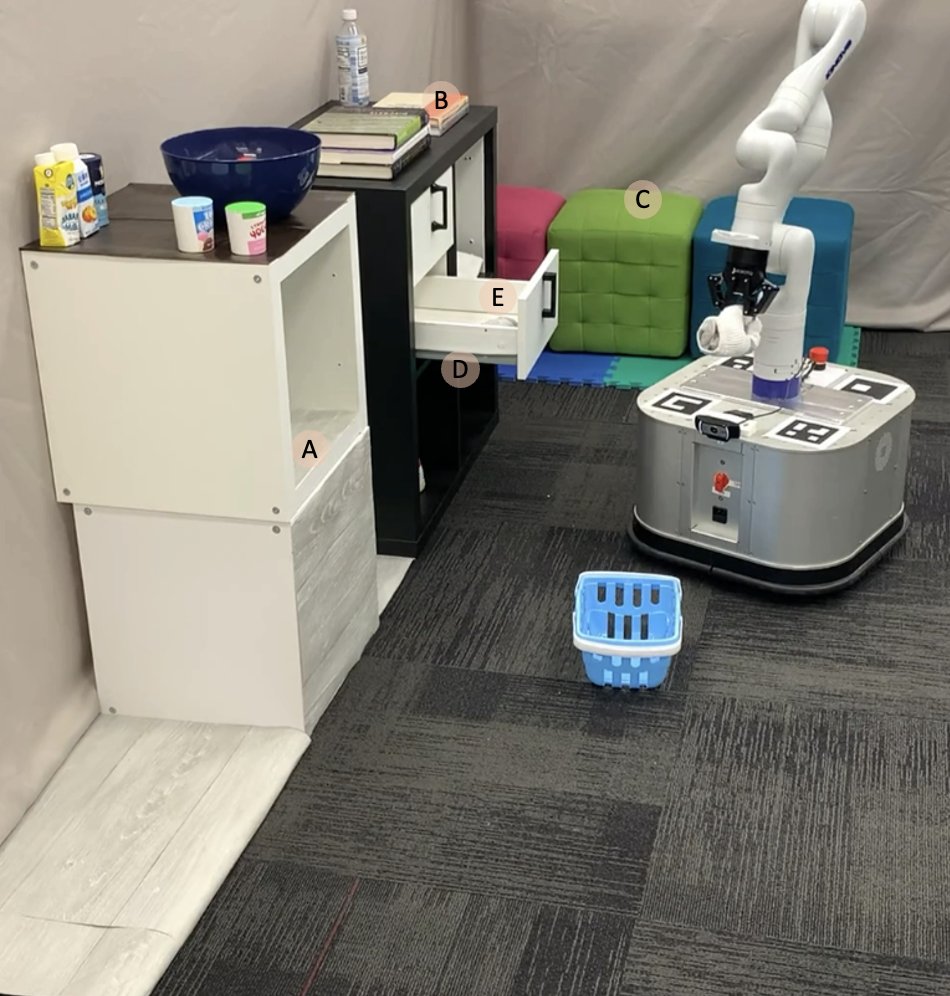}%
{Where should you place the white socks so that they're inside the low drawer? \\\emph{(A) A (B) B (C) C (D) D (E) E}}%
{(E) E}%
{Spatial constraints. Semantic reasoning.}%
{VQA example: Tidybot (scene understanding, geometry spatial reasoning)}



\vqafigure{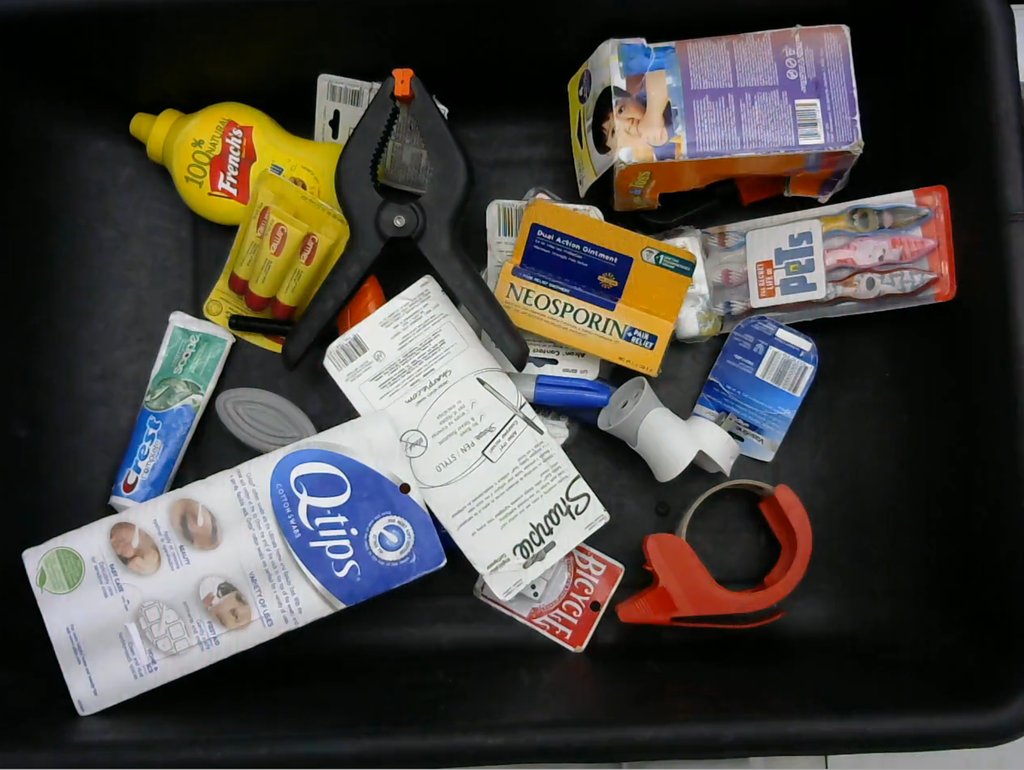}%
{Suppose you have a bimanual robot for bin picking: one arm is equipped with a suction gripper, and the other arm is equipped with a parallel-jaw gripper. The robot first picks up the Sharpie package (white cardboard in the middle). After removing it, which of the following objects is most likely to become possible to grasp? \\\emph{(A) french mustard, take and toss cups, clamp (B) qtips, utility knife, air wick plug (C) sharpie marker, gorilla glue, dice package (D) campbells soup, sardines, bialetti (E) toy gun, lighter, chapstick}}%
{(C) sharpie marker, gorilla glue, dice package}%
{these are originally hidden under sharpie but now visible after removal. originally they only show a tip or small portion}%
{VQA example:  Bin Picking (scene understanding, geometry spatial reasoning)}

\vqafigureduo{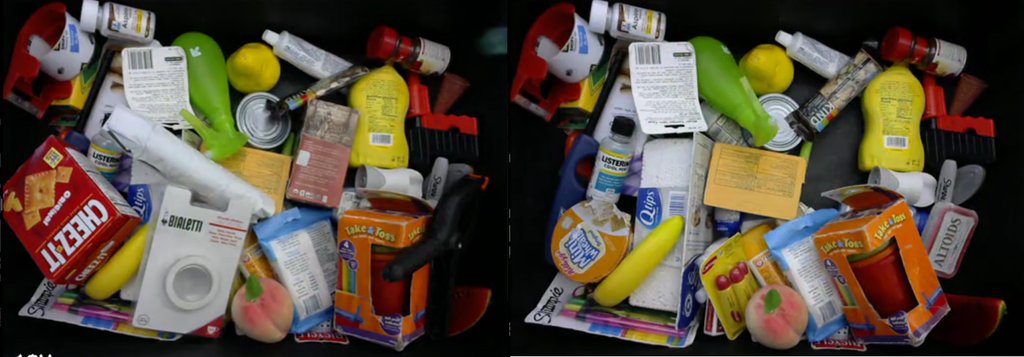}%
{From Image 1 to Image 2, how many items are grasped away successfully \\\emph{(A) 1 (B) 2 (C) 3 (D) 4 (E) 5}}%
{(E) 5}%
{cheezo on the left, tape in the middle, box next to mustard, clamp on the right, the white bar in the middle}%
{VQA example:  Bin Picking (scene understanding, geometry spatial reasoning)}



\vqafigure{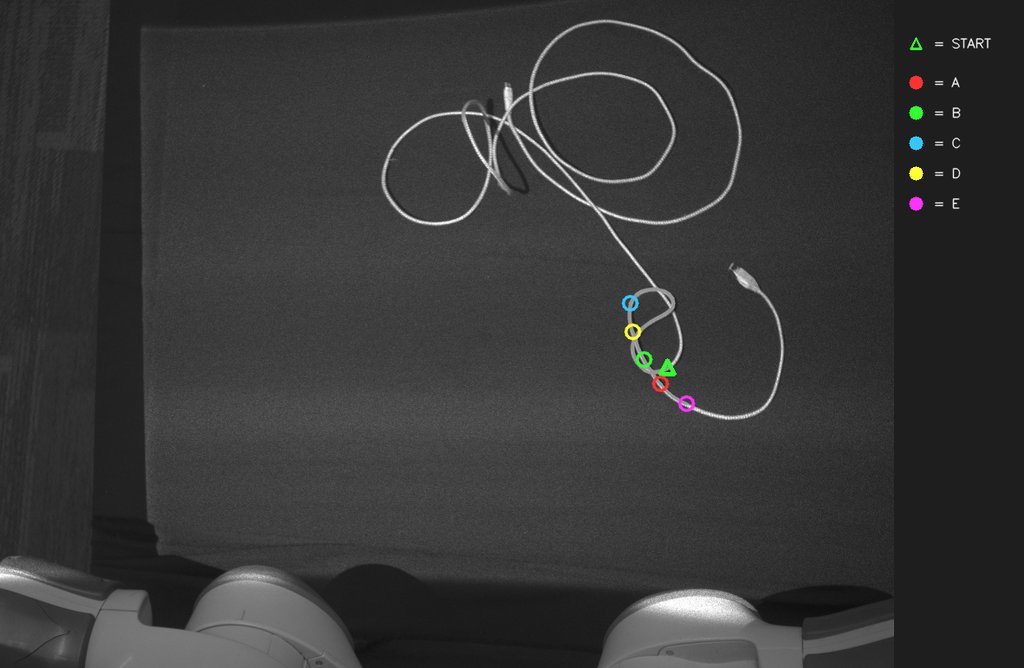}%
{The image shows a cable with a gray traced segment. The green triangle marks START. Five candidate points (A, B, C, D, E) are marked along the trace. Following the trace from START, which point marks the END of this traced segment? \\\emph{(A) Point A (B) Point B (C) Point C (D) Point D (E) Point E}}%
{(E) Point E}%
{Following the gray traced segment from the START marker, Point E marks where the trace ends.}%
{VQA example: Cable Tracing (scene understanding, geometry spatial reasoning)}


\vqafigureduo{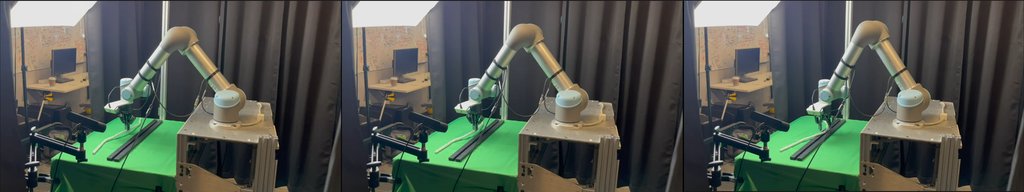}%
{The robot's task is to position the cable completely in the channel. Given the sequence of images, which of these actions is the robot most likely executing? \\\emph{(A) Pick (B) Press (C) Place (D) Slide (E) Return Home}}%
{(C) Place}%
{In the first image, the robot is holding a part of the cable. In the second image, the robot is still holding the cable but is closer to the channel. In the third image, a portion of the cable is on the channel and the robot's gripper is open. Taking this sequence together and in order, the action executed is 'place'. Pick is already executed before the first image and is incompatible with the last image in the order. At no point in the sequence is the gripper applying a downward force that causes the cable to contact the table/channel surface so Press is incorrect. Slide and Return Home are incorrect as well.}%
{VQA example: Gasket Assembly (low level motion awareness, motion feasibility collision avoidance)}


\vqafigure{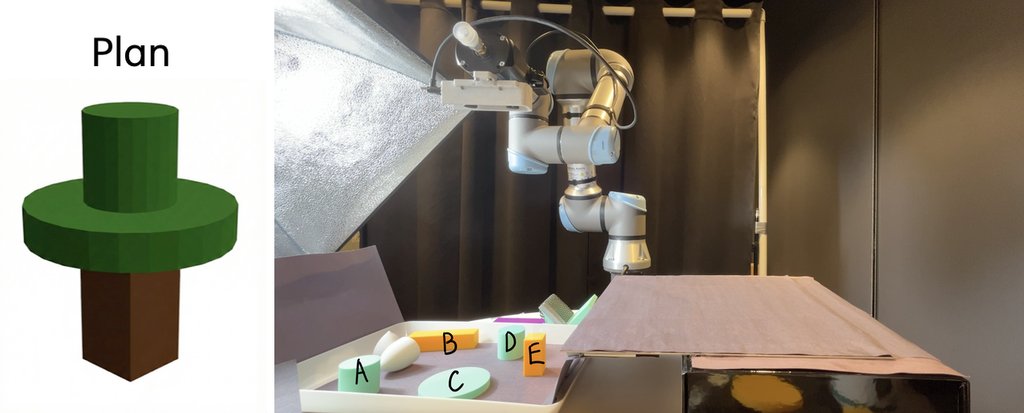}%
{The robot is trying to assemble a structure using only the blocks available in the tray according to a plan illustrated in the first image. Analyze the workspace, which block should the robot pick up and place first? \\\emph{(A) A (B) B (C) C (D) D (E) E}}%
{(E) E}%
{According to the plan, the short cube block should go on the bottom, so it should be picked first.}%
{VQA example: Assembly (high level decision making, goal condition action reasoning)}

\vqafigure{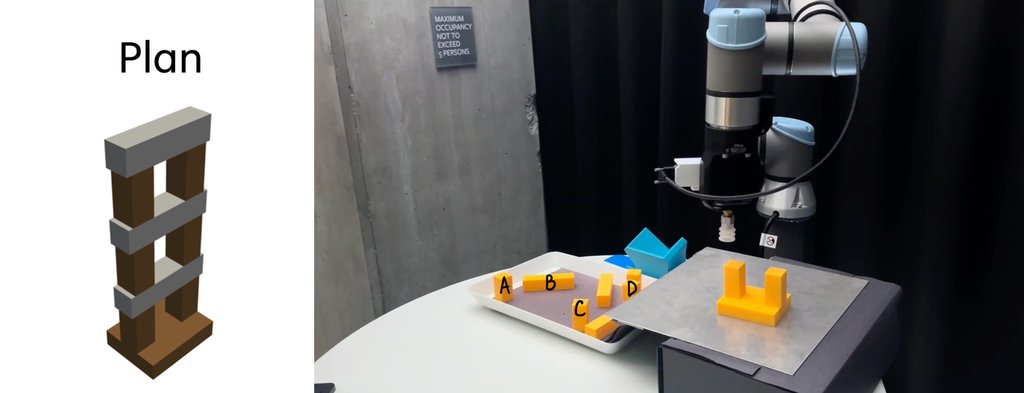}%
{The robot is trying to assemble a structure using only the blocks available in the tray according to a plan illustrated in the first image. Which of the labeled block should it grasp and place first? \\\emph{(A) Block A (B) Block B (C) Block C (D) Block D (E) None of the above}}%
{(B) Block B}%
{Following the plan, the next piece in the structure should be a long square piece. Of the labelled blocks, only Block B fits this description. It should be placed next.}%
{VQA example: Assembly (high level decision making, goal condition action reasoning)}


\vqafigure{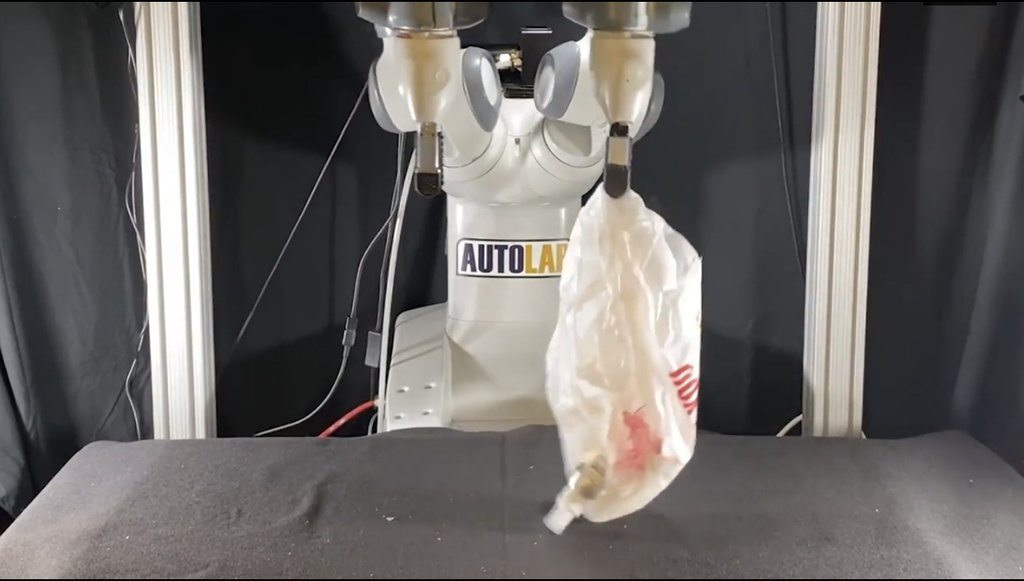}%
{What is the failure reason for this robot bagging? \\\emph{(A) Bag in irrecoverable position (B) Objects placed outside opening (C) Slipped grasp during lifting (D) Objects fall out when lifting (E) None of the above}}%
{(C) Slipped grasp during lifting}%
{The object is in the bag. The grasp failed because one gripper released the bag handle.}%
{VQA example: Bagging (recovery replanning robustness, failure diagnosis)}

\vqafigure{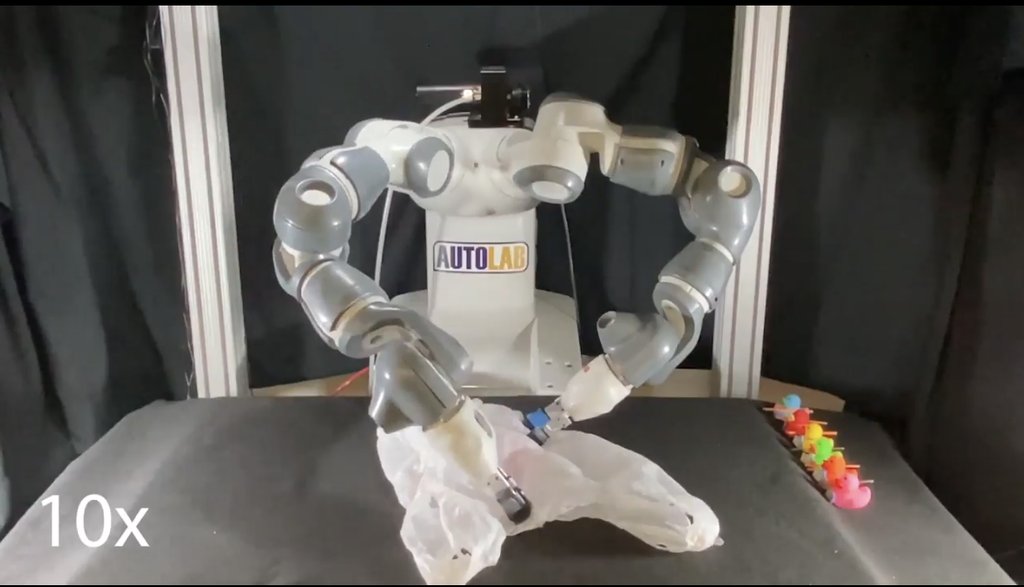}%
{The robot is performing a bagging task, which requires placing all objects into the bag. Given the bag’s current opening direction, in which direction should the robot move the bag so that the opening is closest to the object? \\\emph{(A) Clockwise 180 degrees (B) Clockwise 90 degrees (C) Counter-clockwise 180 degrees (D) Counter-clockwise 90 degrees (E) None of the above}}%
{(D) Counter-clockwise 90 degrees}%
{The bag opening is currently facing the bottom edge of the image, so it needs to rotate toward the left side of the image.}%
{VQA example: Bagging (scene understanding, geometry spatial reasoning)}


\vqafigure{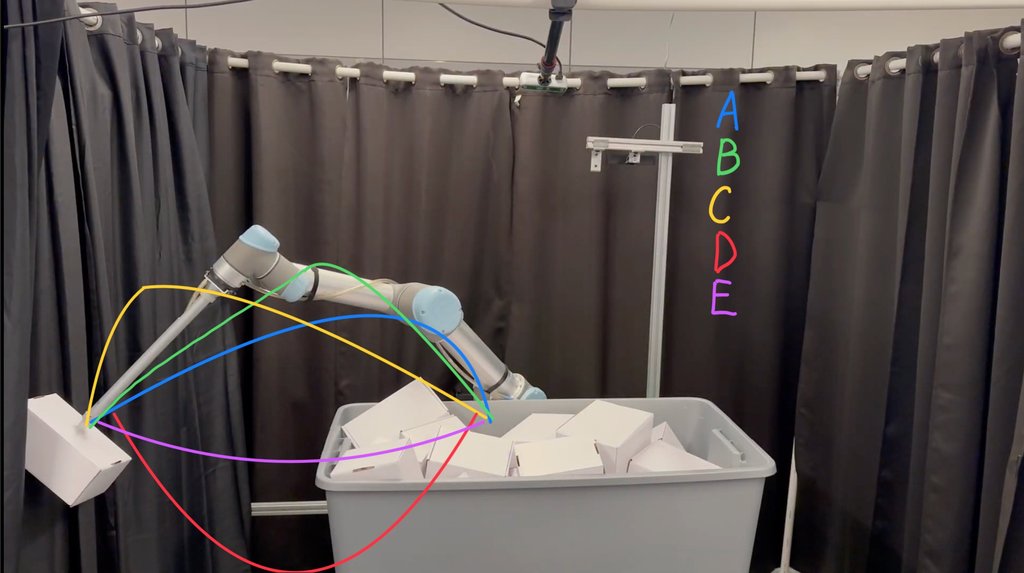}%
{The robot is currently at its goal position after transporting a small box out of the bin. Each letter is color coded on the image, with the same colored curve matching the trajectory that letter represents. Which trajectory is optimal for transporting this box? \\\emph{(A) A (B) B (C) C (D) D (E) E}}%
{(E) E}%
{E's trajectory is faster and could sidesteps the box that is poking out of the box. A's trajectory is valid but is slightly longer. All other trajectories are longer than E and A.}%
{VQA example: Bin Picking (low level motion awareness, motion feasibility collision avoidance)}

\vqafigure{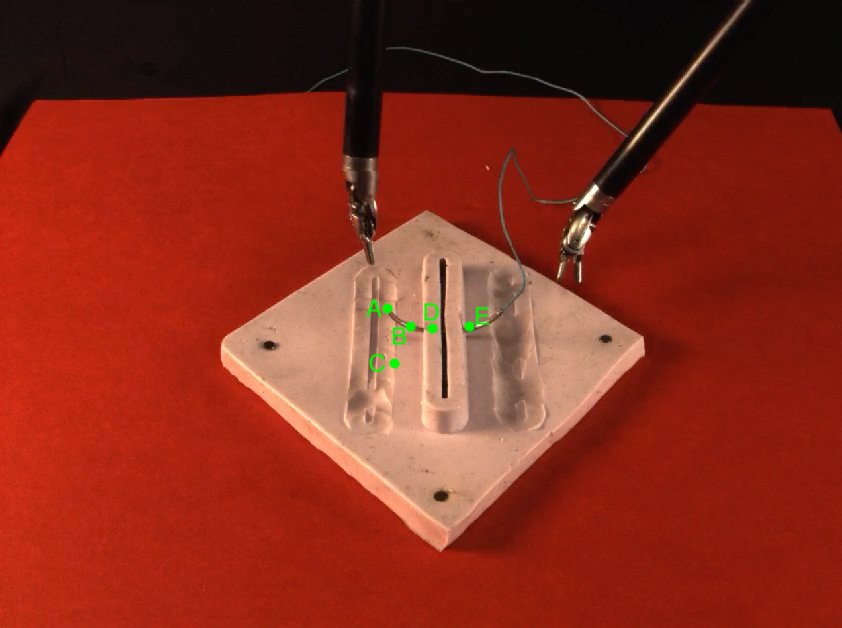}%
{To initiate the pull-through phase, the left gripper needs to extract the needle. Where is the optimal grasping point? \\\emph{(A) Point A (B) Point B (C) Point C (D) Point D (E) Point E}}%
{(B) Point B}%
{Grasping the needle body proximal to the apex avoids tool damage to the sharp tip and provides the necessary leverage for a curvilinear extraction.}%
{VQA example: Surgical Robotics (scene understanding, geometry spatial reasoning)}

\vqafigure{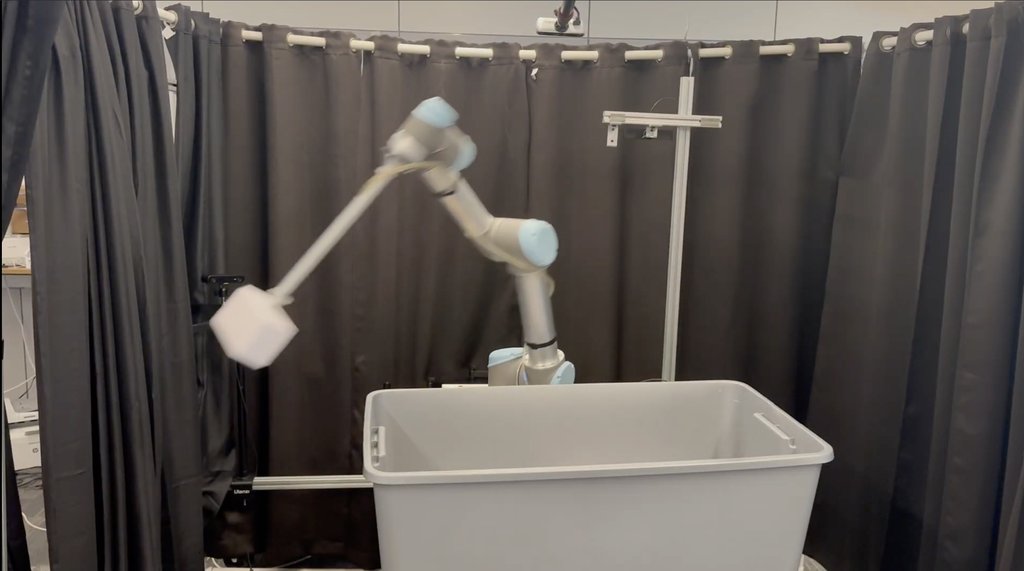}%
{The robot is transporting a white box using a long suction tool to a position on the left of the image. Which of the following is the most accurate analysis of this frame? \\\emph{(A) The robot is successfully transporting the box, and the motion blur indicates high-speed stable transport. (B) The robot has dropped the box due to jerky movements. (C) The suction tool is pressing down on the pile of boxes to compact them into the bin to make space. (D) The robot is hovering the box in a stationary position while the camera moves, creating a blurring effect on the foreground. (E) The robot is executing a dynamic "toss" primitive to throw the object into the back of the bin intentionally.}}%
{(A) The robot is successfully transporting the box, and the motion blur indicates high-speed stable transport.}%
{The box is firmly attached to the robot tool, and is being transported by the robot. Therefore A is correct.}%
{VQA example: Bin Picking (low level motion awareness, motion feasibility collision avoidance)}

\vqafigure{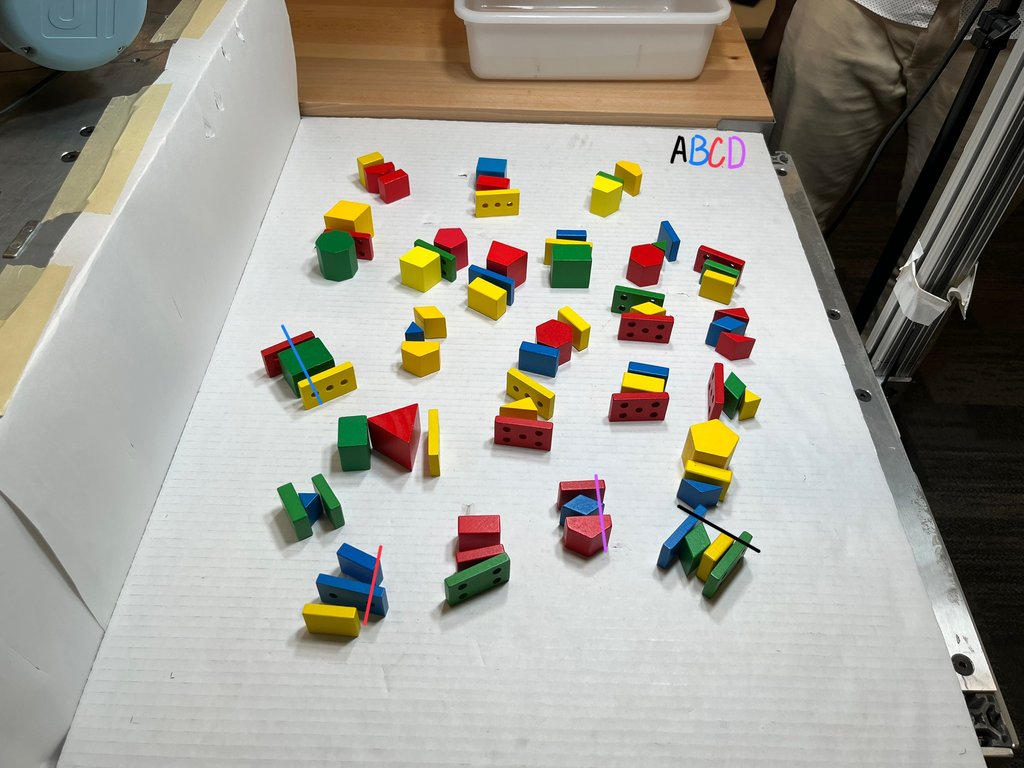}%
{Consider a problem in which a robot attempts to pick up multiple blocks. Analyze the workspace in this image, the lines overlaid on the blocks represent a line draw through the two jaws of a parallel-jaw gripper. The robot will first descend to an appropriate height, and close the gripper in the orientation specified by the line. Which of the following grasps will pick up the most blocks? \\\emph{(A) A (B) B (C) C (D) D (E) They will all pick up the same number of blocks.}}%
{(B) B}%
{Grasp B will cause the 3 cubes to press against each other and can be picked up. The other grasps will cause at least 1 block to slide away from the grasp due to the block geometry, resulting in at most 2 blocks being picked up.}%
{VQA example: Multi-Object Grasping (low level motion awareness, motion feasibility collision avoidance)}


\vqafigure{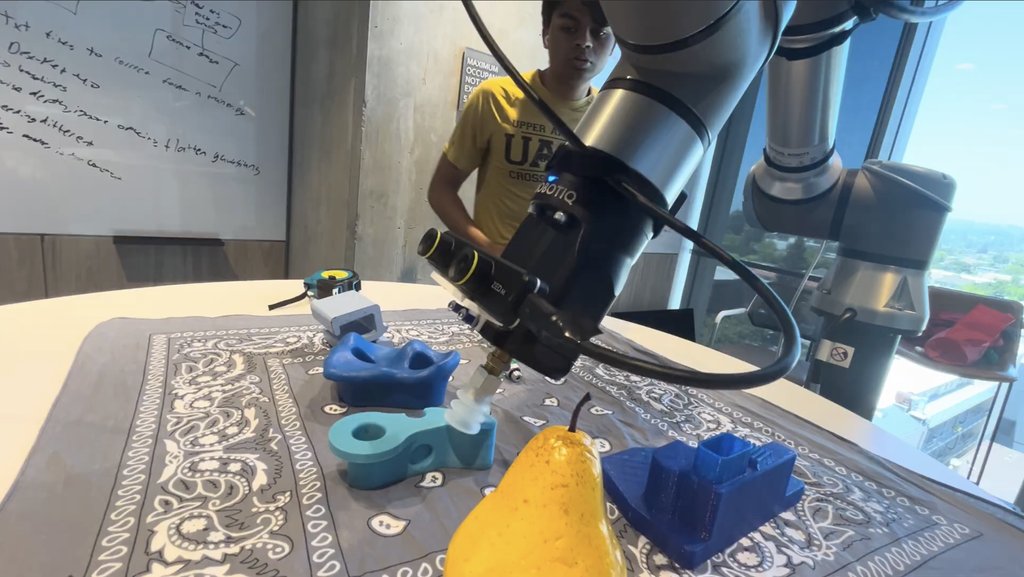}%
{The image above shows the robot attempting a suction grasp. Analyze the grasp position and orientation. Which of the following is correct? \\\emph{(A) The robot will successfully form a suction seal and can lift the object. (B) The robot will successfully form a seal but cannot lift the object because it will collide with an obstacle. (C) The robot will not form a successful seal due to a bad approach angle or position. (D) The gripper has uncompressed bellows, indicating the robot has not yet initiated the grasp attempt. (E) Not enough information.}}%
{(C) The robot will not form a successful seal due to a bad approach angle or position.}%
{The robot is approaching a flat surface at an angle, which is suboptimal. The image also shows that a part of the gripper bellows are not in contact with the surface, which is required to form a suction seal.}%
{VQA example: Suction Grasping (scene understanding, physical interaction)}

\vqafigure{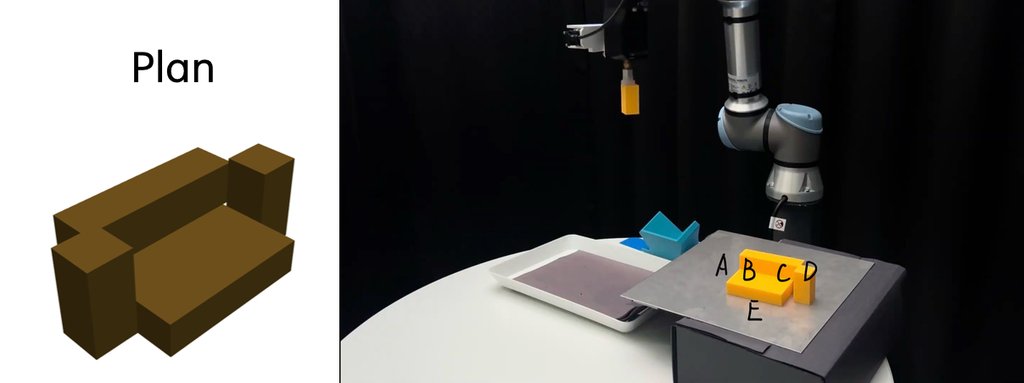}%
{The robot is trying to assemble a structure using only the blocks available in the tray according to a plan illustrated in the first image. Analyze the block the robot current has in its suction gripper, which letter is closest to the location the block should be placed at? \\\emph{(A) A (B) B (C) C (D) D (E) E}}%
{(A) A}%
{According to the plan, the structure should have a short piece on either side. The piece on the right has already been placed, so the next piece should go on the left, where A is.}%
{VQA example: Suction Grasping (high level decision making, goal condition action reasoning)}



\vqafigure{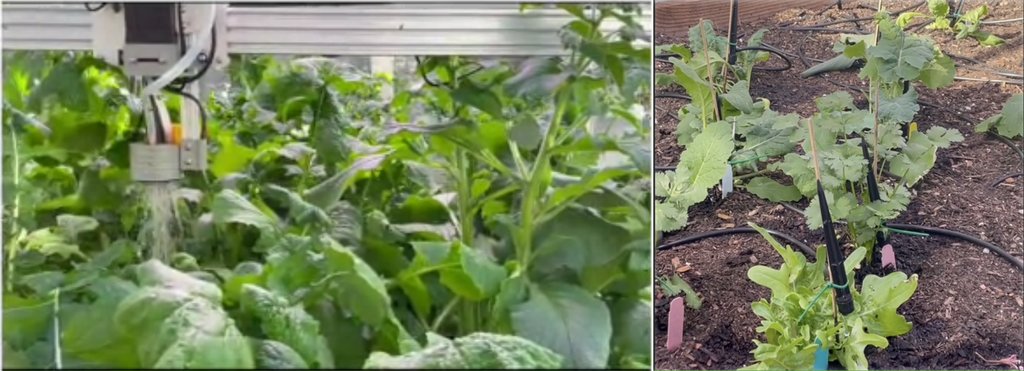}%
{What types of irrigation is being done in the images above? \\\emph{(A) Image 1: Spray irrigation from robot nozzle, Image 2: Drip Irrigation from Emitters (B) Image 1: Drip Irrigation from Emitters, Image 2: Spray irrigation from robot nozzle (C) Both images are showing drip irrigation. (D) Both images are showing spray irrigation. (E) It is impossible to tell what type of irrigation.}}%
{(A) Image 1: Spray irrigation from robot nozzle, Image 2: Drip Irrigation from Emitters}%
{Image 1 shows water flowing from a nozzle while Image 2 shows water flowing from multiple Shrubbler Drip Emitters. Option A is the correct answer.}%
{VQA example: Alphagarden (scene understanding, geometry spatial reasoning)}

\vqafigure{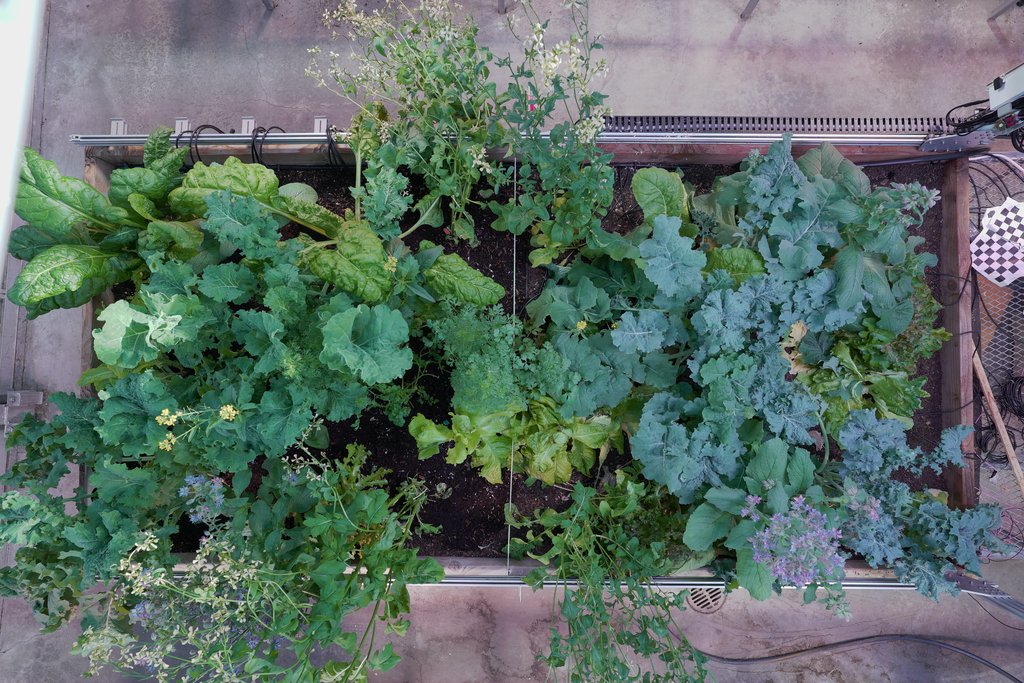}%
{The image shows a top-down view of two gardens separated by a white line. As a measure of growth, which of the following is most likely to be correct? \\\emph{(A) While both gardens are not using most of their allotted garden beds space, the left garden uses more of its allotted garden bed space. (B) While both gardens are not using most of their allotted garden beds space, the right garden uses more of its allotted garden bed space. (C) It is not possible to tell which garden uses more of its garden bed space. (D) Both gardens are using most of their allotted garden beds space, and the right garden uses more of its allotted garden bed space. (E) Both gardens are using most of their allotted garden beds space, and the left garden uses more of its allotted garden bed space.}}%
{(D) Both gardens are using most of their allotted garden beds space, and the right garden uses more of its allotted garden bed space.}%
{From a strictly top-down view, considering canopy cover, both gardens use most of their garden bed space. However, the right garden uses more of its allotted garden bed space. Option D is the correct answer.}%
{VQA example: Alphagarden (scene understanding, geometry spatial reasoning)}


\vqafigure{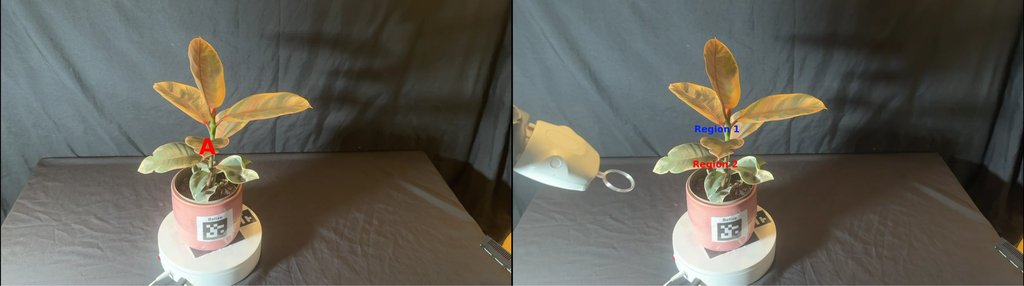}%
{The robot's task is to push down leaf A from image 1. In image 2, the robot is approaching the leaf. Which of the following action sequences is correct? \\\emph{(A) The robot starts in Region 1 and moves upward to contact the leaf. (B) The robot starts in Region 1 and moves downward to contact the leaf. (C) The robot starts in Region 2 and moves upward to contact the leaf. (D) The robot starts in Region 2 and moves downward to contact the leaf. (E) There is no way to achieve the task given the current plant position.}}%
{(B) The robot starts in Region 1 and moves downward to contact the leaf.}%
{Pushing leaf A down requires the robot to start in Region 1. So options C and D are wrong. Next, the robot must move downward. So A is wrong. Option B is the correct answer."}%
{VQA example: Botany-Bot (low level motion awareness, motion feasibility collision avoidance)}

\vqafigureduo{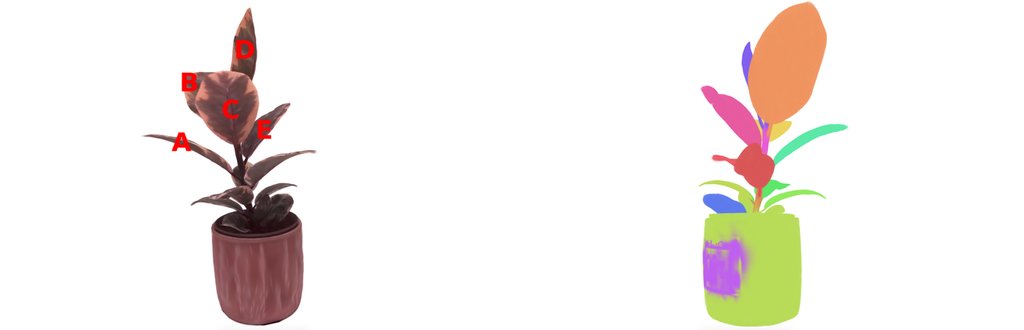}%
{The plant in Image 1 is shown with its parts segmented in Image 2. Additionally, the view in Image 2 is rotated. Match the labeled leaves to their corresponding leaves in Image 2 using the segmentation colors. \\\emph{(A) A: Green, B: Yellow, C: Magenta, D:  Orange, E: Violet (B) A: Orange, B: Violet,  C:  Green, D:  Yellow,  E: Magenta (C) A: Green, B: Orange C: Yellow, D: Violet, E: Magenta (D) A: Yellow,  B: Magenta, C: Orange, D: Violet, E: Green (E) A: Violet, B: Green, C: Yellow, D: Magenta, E: Orange}}%
{(C) A: Green, B: Orange C: Yellow, D: Violet, E: Magenta}%
{Based on the shape and size of the leaves, and the plant structure,  we are able to determine the corresponding colors.  Option C is the correct answer}%
{VQA example: Botany-Bot (scene understanding, geometry spatial reasoning)}


\vqafigureduo{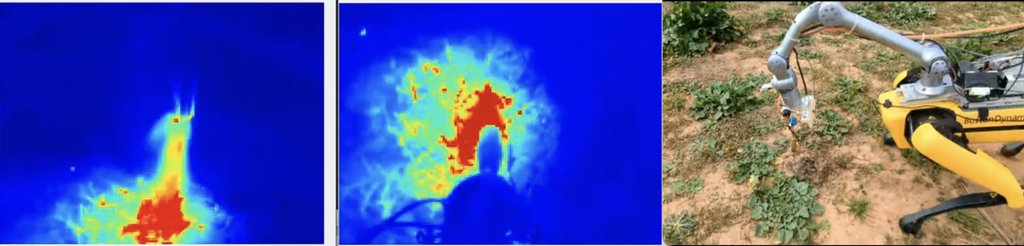}%
{The robot’s current position as it holds a flaming torch close to the ground for weed removal. The system uses two thermal cameras: one mounted on the side to capture a lateral view of the flame, and another pointing downward to observe the flame’s thermal footprint on the ground. Based on the flame’s shape and motion, determine whether wind is present in the field and whether it is affecting the flame. \\\emph{(A) The wind is blowing toward the left side of the thermal images. (B) There is no wind. (C) The wind is blowing toward the right side of the thermal images. (D) The wind is blowing upward in the thermal images (from bottom to top). (E) The wind is blowing downward in the thermal images (from top to bottom).}}%
{(A) The wind is blowing toward the left side of the thermal images.}%
{The flame’s thermal footprint on the ground indicates that the heat is displaced toward the left side of the image. The lateral thermal view also shows the flame bending toward the left.}%
{VQA example: Weed Removal (scene understanding, geometry spatial reasoning)}

\vqafigure{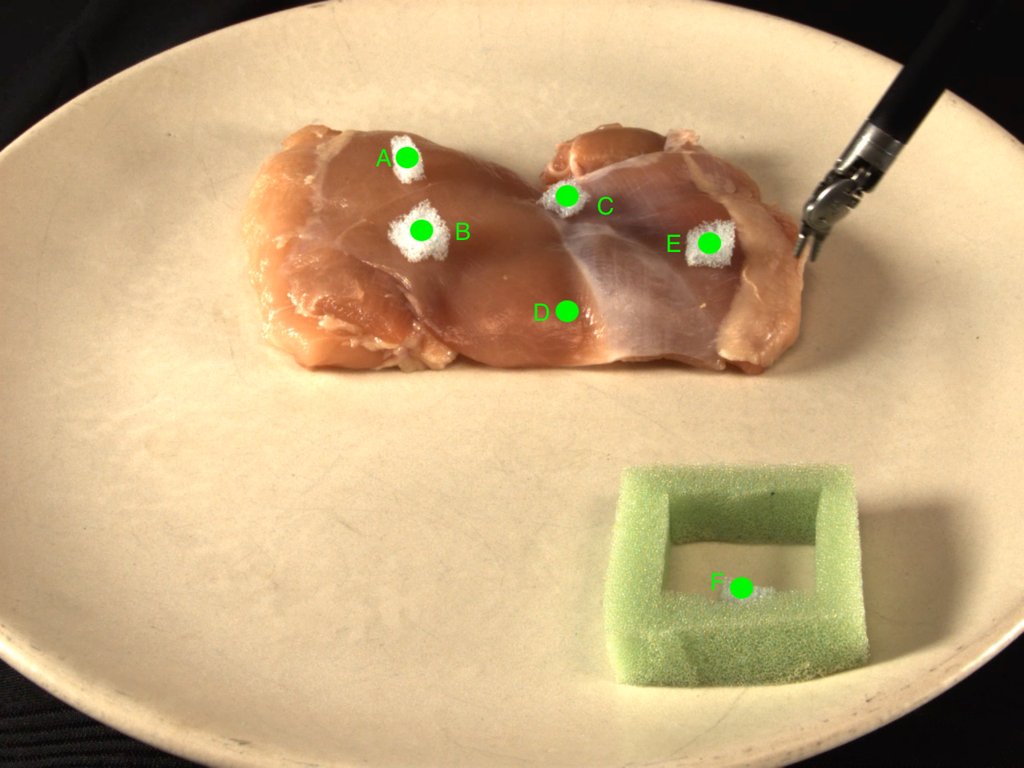}%
{Where are the white foam fragments that remain on the surface of the chicken flesh? \\\emph{(A) (A, B, C, D) (B) (A, B, C, E) (C) (A, B, D, E) (D) (B, C, D, E) (E) (A, B, C, F)}}%
{(B) (A, B, C, E)}%
{Four distinct white foam blocks remain on the chicken flesh surface. One additional piece is already in the collection box.}%
{VQA example: Surgical Robotics (scene understanding, geometry spatial reasoning)}






